\pdfoutput=1
\documentclass[preprint]{acmart}
\usepackage{forest}
\usetikzlibrary{angles}
\usepackage{multicol}
\usepackage{tcolorbox}
\usepackage{svg}
\usepackage{tcolorbox}
\usepackage{svg}
\usepackage{mathtools}
\usepackage{amssymb,amsmath}
\usepackage{multirow}
\usepackage{graphicx}
\usepackage{subfig}

\def\BibTeX{{\rm B\kern-.05em{\sc i\kern-.025em b}\kern-.08emT\kern-.1667em\lower.7ex\hbox{E}\kern-.125emX}}

\copyrightyear{}
\acmYear{}
\setcopyright{none}
\acmConference{}
\acmBooktitle{}
\acmPrice{}
\acmDOI{}
\acmISBN{}

\newcommand{\argmin}{\mathop{\mathrm{argmin}}}

\begin{document}

%
\title{Test Selection for Deep Learning Systems}

%

\author{Wei Ma}
\affiliation{\institution{SnT, University of Luxembourg}}
\email{wei.ma@uni.lu}

\author{Mike Papadakis}
\affiliation{\institution{SnT, University of Luxembourg}}
\email{michail.papadakis@uni.lu}

\author{Anestis Tsakmalis}
\affiliation{\institution{SnT, University of Luxembourg}}
\email{letsakkiller@gmail.com}

\author{Maxime Cordy}
\affiliation{\institution{SnT, University of Luxembourg}}
\email{maxime.cordy@uni.lu}
 
\author{Yves Le Traon}
\affiliation{\institution{SnT, University of Luxembourg}}
\email{yves.letraon@uni.lu}

%
\renewcommand{\shortauthors}{Ma, et al.}

%
%
\begin{abstract}
Testing of deep learning models is challenging due to the excessive number and complexity of computations involved. As a result, test data selection is performed manually and in an ad hoc way. This raises the question of how we can automatically select candidate test data to test deep learning models. Recent research has focused on adapting test selection metrics from code-based software testing (such as coverage) to deep learning. However, deep learning models have different attributes from code such as spread of computations across the entire network reflecting training data properties, balance of neuron weights and redundancy (use of many more neurons than needed). Such differences make code-based metrics inappropriate to select data that can challenge the models (can trigger misclassification). We thus propose a set of test selection metrics based on the notion of model uncertainty (model confidence on specific inputs). Intuitively, the more uncertain we are about a candidate sample, the more likely it is that this sample triggers a misclassification. Similarly, the samples for which we are the most uncertain, are the most informative and should be used to improve the model by retraining. We evaluate these metrics on two widely-used image classification problems involving real and artificial (adversarial) data. We show that uncertainty-based metrics have a strong ability to select data that are misclassified and lead to major improvement in classification accuracy during retraining: up to 80\% more gain than random selection and other state-of-the-art metrics on one dataset and up to 29\% on the other. 

\end{abstract}


%

%
\maketitle

\section{Introduction}

Deep Learning (DL) systems \cite{LeCunBH15} are capable of solving complex tasks, in many cases equally well or even better than humans. Such systems are attractive because they learn by themselves, i.e., they require only minimum human knowledge. This property makes DL simple, flexible and powerful. As a result, it is increasingly used and integrated with larger software systems and applications. 

Naturally, the adoption of this technology introduces the need for its reliable assessment. In classical, code-based software engineering, this assessment is realised by means of testing. However, the test of DL systems is challenging due to the complexity of the tasks they solve. In order to effectively test the DL system, we need to identify corner cases that challenge the learned properties. In essence, DL system testing should focus on identifying the incorrectly learned properties and lead to data that can make the systems deviate from their expected behaviour. 

To this end, recent research \cite{ma2018deepgauge, Pei2017, Kim2019, TianPJR18} focuses on adapting test selection metrics from code-based software testing (such as code coverage) to DL systems. However, such code metrics assume that incorrectly learned properties reside on particular instances of the program's control and data flow, which is not the case of DL. The behaviour of DL systems depends on all the computations across the entire network, which in the case of well-trained models are balanced (e.g. single neuron activations only have minor influence) \cite{SrivastavaHKSS14}. Perhaps more importantly, DL models reflect particular properties of the training data and their behaviour is determined based on the knowledge they acquired during the training phase. Such differences make code-based metrics insufficient to select data that can effectively challenge the models (i.e. can trigger misclassification) \cite{Kim2019}.

Of course automated test selection techniques can play an important role when considering different scenarios/assumptions and can be considered as a first step towards automating the manual and non-systematic test selection that is currently performed. However, such techniques have an important limitation: they are not immediately linked to test effectiveness \cite{Kim2019}. Thus, the causal relationship between test selection and triggered misclassifications\footnote{`faults' according to the traditional software testing terminology} is weak. Moreover, practitioners using such metrics can neither measure the added value of the selected test samples (improved confidence) nor improve the quality (e.g. accuracy) of the system. According to Kim et al. ``for a test adequacy criterion to be practically useful, it should be able to guide the selection of individual inputs, eventually resulting in improvements of the accuracy of the DL system under investigation'' \cite{Kim2019}. 

Experience has shown that classification mistakes are incorrectly learned properties that happen due to overlapping and closely located regions of the feature space.  Therefore, the cases residing between the learned categories and their boundaries are the most likely to be the incorrectly learned ones. In view of this, rather than aiming for the coverage of specific neurons \cite{Pei2017} or test data diversity \cite{Kim2019}, we argue that testing should focus on data that have properties close to the model boundaries. In other words one should direct test selection towards the boundaries of the learned classes. 

To make test selection practical, we propose a set of test selection metrics (called adequacy metrics by Kim et al. \cite{Kim2019}) for DL systems. We define such metrics based on the notion of model uncertainty (low confidence on specific inputs). Intuitively, the more uncertain a model is about a candidate sample, the more likely it can make a misclassification. Similarly, the samples for which the model is the most uncertain are expected to be the most informative and to maximize the information that models can learn from (should be used for model improvement - retraining). We approximate uncertainty using the variance caused by multiple dropouts \cite{SrivastavaHKSS14}, dropping arbitrary neurons (along with their connections) from the network and observing the impact on its predictions, as suggested by Gal and Ghahramani \cite{gal2016dropout}. We also use the actual model's output probabilities as a certainty measure, which we can also combine with dropout variance.

We evaluate these metrics using image classifiers on two widely used datasets.  We examine the performance of the metrics with respect to the current state of the art, i.e. the surprise adequacy metrics introduced by Kim et al. \cite{Kim2019}. In particular, we investigate the correlation between the metrics and misclassification on both real and artificial (adversarial) data. We show that uncertainty-based metrics have medium to strong correlations with misclassification when considering real data, and strong correlations when considering a mix of real and adversarial data. We also show that prediction probabilities (a simple certainty metric overlooked by previous work) is among the most effective metrics, significantly outperforming the state of the art. Finally, we show that a combination of the dropout variance with prediction probabilities can lead to major improvements in classification accuracy (when selecting data for retraining) compared to surprise adequacy and random selection.

Our contributions can be summarised by the following points: 

\begin{itemize}
\item We propose a set of test selection metrics based on the notion of model uncertainty, i.e., the confidence in classifying correctly unseen inputs. We consider the variance caused by multiple dropouts (i.e. the distribution of the model's output under dropouts), the model's prediction probabilities, and metrics combining the two. 

\item We demonstrate that the uncertainty-based metrics significantly challenge DL models and have moderate to strong correlations with misclassification (correlations of approximately 0.4 on real data and 0.6 on real plus adversarial ones). Furthermore, our metrics significantly outperform the current state of the art (the surprise adequacy metrics). 

\item We also show that model uncertainty can guide the selection of informative input data, i.e., data that are capable of increasing classification accuracy. In particular, when retraining the DL model based on the selected data, our best performing metrics achieve up to 80\% improvement over random selection and surprise adequacy on one dataset, and up to 29\% on the other. 
 
\end{itemize}

%
\section{Related Work}

Testing of learning systems is typically performed by selecting a dedicated test set randomly from available labelled data~\cite{Witten2011}. When an explicit test dataset is not available, one can rely on cross-validation~\cite{Kohavi1995} to use part of the training set to anticipate how well the learning model will generalize to new data. These established procedures, however, often fail to cover many errors. For instance, research work in adversarial learning has shown that applying minor perturbations to data can make models give a wrong answer~\cite{Goodfellow2015}. Nowadays, those adversarial samples remain hard to detect and bypass many state-of-the-art detection methods~\cite{Carlini2017}. In order to achieve better testing, multiple approaches have been proposed in the recent years. We distinguish four categories of contributions: (i) metrics for measuring the coverage of a test set; (ii) generation of artificial inputs; (iii) metrics for selecting test data; (iv) detection of adversarial samples.

DeepXplore, proposed by  Pei et al.~\cite{Pei2017}, comprises both a coverage metric and a new input generation method. It introduces neuron coverage as the first white-box metric to assess how well a test set covers the logic of DL models. Leaning on this criterion, DeepXplore generates artificial inputs by solving a joint optimization problem with two objectives: maximizing the behavioural differences between multiple models and maximizing the number of activated neurons. Pei et al. report that DeepXplore is effective at revealing errors (misclassifications) that were undetected by conventional ML evaluation methods. Also, retraining with additional data generated by DeepXplore increases the accuracy of the models. On some models, the increase is superior (1 to 3\%) to an increase obtained by retraining with data generated by some adversarial technique~\cite{Goodfellow2015}. Pei et al. also show that randomly-selected test data and adversarial data achieve smaller neuron coverage than data generated by DeepXplore. While they assume that more neuron coverage leads to better testing, future research showed that this metric is inadequate~\cite{Ma2018,Kim2019}.

In a follow-up paper, Tian et al.~\cite{Tian2018} propose DeepTest as another method to generate new inputs for autonomous driving DL models. They leverage metamorphic relations that hold in this specific context. Like DeepXplore, DeepTest utilizes the assumption that maximising neuron coverage leads to more challenging test data. The authors show that different image transformations lead to different neuron coverage values and infer that neuron coverage is an adequate metric to drive the generation of challenging test data. However, this claim was not supported by empirical evidence.

DeepGauge~\cite{Ma2018} and DeepMutation~\cite{Ma2018b} are two test coverage metrics proposed by Ma et al. With DeepGauge, they push further the idea that higher coverage of the structure of DL models is a good indicator of the effectiveness of test data. They show, however, that the basic neuron coverage proposed previously is unable to differentiate adversarial attacks from legit test data, which tends to indicate the inadequacy of this metric. As a result, they propose alternative criteria with different granularities, i.e. at the neuron level and the layer level. Their experiments reveal that replacing original test inputs by adversarial ones increase the coverage of the model wrt. DeepGauge's criteria. However, they never assessed the capability of their criteria to guide test selection and retraining.

Similarly, DeepMutation leverages the mutation score used in traditional mutation testing to DL models. From a given model, it generates multiple mutant models by applying different operators such as, e.g., neuron switch or layer removal. Then, they define the mutation score of a test input as the number of mutants that it killed (i.e. those that yield a different classification output for the test input than the original model). The fundamental difference with DeepGauge lies in that mutation score rather assesses how sensitive the model is wrt. to the test inputs. 

Nevertheless, both DeepGauge and DeepMutation settle for indicative metrics that were not used to guide test selection. Moreover, a recent study~\cite{Kim2019} has shown that neuron coverage criteria do not necessarily increase when more misclassified/surprising inputs are added. Although our contribution starts from the similar idea of analyzing the sensitivity of the model by mutating it (using dropouts), our scope differs from DeepMutation in that we examine how mutations can actually support the selection of data for testing and improving (retraining) the model.

Later on, Ma et al.~\cite{Ma2019}, inspired from combinatorial interaction testing, propose DeepCT: a test coverage metric and a generation method based on $t$-wise coverage criteria. Within a given layer, all $t$-uples of neurons should be covered by at least one test input. They also propose an algorithm to generate artificial inputs to cover as many t-wise interactions as possible. In their evaluation, they show, first, that random test selection cannot cover a large part ($>$ 65\%) of the 2-wise neuron interactions. Second, they show that retraining on the inputs generated by their algorithm allows the detection of up to $10\%$ of adversarial samples that could not be detected by retraining on randomly selected inputs. An alternative proposed by Xiaofei Xie et al.~\cite{Xie2018} is DeepHunter, a fuzzing-based test generation algorithm to hunt defects in DL models. The fuzzing is guided by the coverage metrics defined in DeepXplore~\cite{Pei2017} and DeepGauge~\cite{Ma2018}. Their evaluation shows that the fuzzing algorithm indeed manages to increase the intended coverage metrics, and is able to discriminate models with different performance. Both DeepCT and DeepHunter focus on generating artificial inputs and are not directed towards the better selection of challenging data for testing and retraining.

Most recently, Kim et al.~\cite{Kim2019} were the first to propose techniques that act as metrics for test coverage \emph{and} test data selection. They highlight the fact that the previous coverage criteria fail to discriminate the added value of \emph{individual} test inputs and are therefore impractical when it comes to selecting appropriate test data. They argue that a test adequacy criterion should guide the selection of individual inputs and eventually help to improve the performance of the DL model. As a consequence, they propose a new metric named \emph{surprise adequacy}, which measures how surprised the model is when confronted with a new input. More precisely, the degree of surprise measures the dissimilarity of the neurons' activation values when confronted with this new input wrt. the neurons' activation values when confronted to the training data. Then, they hypothesise that a set of test inputs is more adequate for both testing and retraining when it covers a more diverse range of surprise values. In other words, a good data set should range from very similar to very different inputs to those observed during training. Kim et al. show experimentally that (a) surprise coverage is sensitive to adversarial samples and (b) retraining on such samples yields better improvements as those cover a wider range of surprise. In this paper, we show that by focusing on model uncertainty rather than input diversity, one can select challenging inputs that are more effective at triggering misclassification and improving the accuracy of the model. We actually regard surprise adequacy and our work as complementary: surprise adequacy aims for a diversity of surprise degrees and thus better applies to models that are not yet well-trained, while our uncertainty metrics are meant to reinforce well-trained models against classes of inputs that remain challenging.

An akin notion of surprise was used by Feinman et al.~\cite{Feinman2017} to detect adversarial samples. This detection method makes use of Kernel density estimate (similar to likelihood-based surprise adequacy \cite{Kim2019}) and Bayesian uncertainty based on dropout variance (similar to the one used in this paper). Another recent method to detect adversarial samples was proposed by Wang et al.~\cite{Wang2019}. It consists of computing how much the labels predicted by a model change when (after training) this model is slightly mutated. Their experiments show that adversarial inputs are more likely to increase the label change rate. Additionally, a purely Bayesian Generative Adversarial is proposed in \cite{Saatchi2017}, where the adversarial sample generator and the discriminator are Bayesian Neural Networks trained with stochastic gradient Hamiltonian Monte Carlo sampling. More specifically, the discriminator network is capable of efficiently detecting adversarial samples exactly because of the competition-based structure, which forces learning to be a repeated contest between the generator and the discriminator. While this line of work was a source of inspiration and while our own study shows that our metrics are somehow sensitive to adversarial generation processes, our application scope is different and does not encompass the detection of adversarial samples. Moreover, dropout variance is but one of the multiple alternatives that form our technique. 

\section{Motivation}

Our work aims at defining data selection metrics for testing and (re)training of DL models. Our key motivation is on the difficulty and complexity of the task, which is typically performed by manual and non-systematic procedures. We thus aim at answering the following two questions:

\begin{description}
\item[-]
 \emph{How can we select test data to challenge (trigger misclassifications in) a Deep Learning model?}
\end{description}

\begin{description}
\item[-]
 \emph{How can we select additional training data to improve the performance (increase classification accuracy) of a Deep Learning model?}
\end{description}

Our goal is to propose objective and measurable ways of determining the effectiveness of DL systems. Given that practitioners cannot classify (label) all possible inputs, they resort on randomly chosen sets. To this end, we equip them with metrics to sample (and label) the most challenging cases (inputs that can trigger misclassifications) and overall reveal potential corner cases of the learned models. Of course we assume that practitioners have access to large sets of unlabelled data, which require manual inspection and analysis. Minimizing the manual effort involved in those activities forms our target. 

Another use case of our uncertainty measures regards their integration with monitoring and decision-making mechanisms that control and operate on top of DL systems. Using such metrics one can easily assess the certainty/uncertainty of taking a decision based on the recommendation of a learned model. 

We perform our study to investigate the ability of data selection metrics when using alternatively real data, adversarial data, and a mix of real and adversarial data. We perform the study using well-trained models (over 90\% accuracy), which are both hard to challenge and to improve.

\section{Test Selection Metrics}


\subsection{Surprise Adequacy Metrics}

The two test selection methods we retain from the literature are those based on surprise adequacy~\cite{Kim2019}. In their recent paper, Kim et al. proposed two metrics to measure the surprise of a DL model $D$ when confronted to a new input $x$. The first one is based on Kernel density estimation and aims at estimating the relative likelihood of $x$ wrt. the training set in terms of the activation values of $D$'s neurons. To reduce computational cost, only the neurons of a specified layer are considered~\cite{Kim2019}.

\begin{definition}
Let $D$ be a DL model trained on a set $T$ of inputs. The \textbf{Likelihood-based Surprise Adequacy} (LSA) of the input $x$ wrt. $D$ is given by
$$LSA(x) = \frac{1}{|A_{N_L} (T)|} \sum_{x_i \in T} K_H (\alpha_{N_L}(x) - \alpha_{N_L}(x_i))$$
where $\alpha_{N_L}(x')$ is the vector recording the activation values of the neurons in layer $L$ of $D$ when confronted to $x$,  $A_{N_L} (T) = \{\alpha_{N_L}(x_i) \mid x_i \in T\}$, and $K_H$ is the Gaussian kernel function with bandwidth matrix $H$.
\end{definition} 

As an alternative, Kim et al. proposed a second metric that relies on Euclidean distance instead of Kernel density estimate. The idea is that inputs that are closer to inputs of other classes and farther from inputs of their own class are considered as more surprising. This degree of surprise is measured as the quotient between the distance of the closest input $x_a$ of the same class as $x$ and the distance of the closest input $x_b$ from any other class. Like the LSA metric, all these distances are considered in the activation value space of the inputs. 

\begin{definition}
Let $D$ be a DL model trained on a set $T$ of inputs. The \textbf{Distance-based Surprise Adequacy} (DSA) of the input $x$ wrt. $D$ is given by
$$DSA(x) = \frac{||\alpha_N (x) - \alpha_N (x_a)||}{||\alpha_N (x) - \alpha_N (x_b)||}$$
where
\begin{align*}
	x_a &= \argmin_{\{x_i \in X \mid D(x_i) = D(x)\}} ||\alpha_N (x) - \alpha_N (x_i)|| \\
	x_b &= \argmin_{\{x_j \in X \mid D(x_j) \in C \setminus D(x)\}} ||\alpha_N (x) - \alpha_N (x_j)|| \\
\end{align*}
and where $D(x')$ is the predicted class of $x'$ by $D$ and $\alpha_N(x')$ is the activation value vector of all neurons of $D$ when confronted to $x'$.
\end{definition}

\subsection{Model Uncertainty Metrics}

The starting point for our own selection metrics lies in the hypothesis that test inputs are more challenging (i.e. more likely to be misclassified) as they engender more uncertainty (as opposed to surprise) in the considered DL model. 

The prediction probabilities of the classes returned by the model are immediate metrics that can indicate how challenging a particular input is. Indeed, one can intuitively state that more challenging inputs are classified with lower probability, that is, the highest prediction probability output by the model is low.

\begin{definition}
Let $D$ be a trained DL model. The \textbf{maximum probability score} of the input $x$ wrt. $D$ is given by
$$MaxP(x) = \max_{i = 1:C} p_i (x)$$
where $C$ is the number of classes and $p_i (x)$ is the prediction probability of $x$ to class $i$ according to $D$. 
\end{definition}

Recently, it has been mathematically proven that neuron dropout \cite{Srivastava2014} can be used to model uncertainty~\cite{gal2016dropout,Kendall2017WhatUD}. Dropout was initially proposed as a technique to avoid overfitting in neural networks by randomly dropping (i.e. disabling) neurons during training~\cite{Srivastava2014}. It was then adapted to estimate the uncertainty of a trained model wrt. a new input $x$~\cite{gal2016dropout}. More precisely, the uncertainty is estimated by sampling $k$ dropped-out models and computing the variance of their resulting prediction probabilities over $x$. Intuitively, while prediction probabilities can be visualized as the distances from $x$ to the class boundaries, the dropout variance represents the variance of these distances induced by the uncertainty of the actual location of the class boundaries.

Formally, let $D$ be an original, well-trained model. Let $\{M_1 \dots M_k\}$ be a set of $k$ mutated models, such that each $M_j$ results from $D$ after randomly dropping out individual neurons with probability (i.e. dropout rate) $r$. Here, we consider permanent dropout in the sense that a mutated model never recovers its dropped-out neurons. Given an input $x$, we denote by $p_i^j (x)$ the prediction probability of $x$ to class $i$ output by $M_j$. We also denote by $P_i(x) = \{p_i^j (x) | 1 \leq j \leq k\}$ the multiset of prediction probabilities of $x$ assigned to class $i$ by all mutants $M_1$ to $M_k$. Then the variance of $P_i(x)$ is a good estimate of the uncertainty of $D$ when classifying $x$ in class $i$.

Following our hypothesis that uncertain inputs are more likely to be misclassified, we define a metric derived from dropout variance to assess how much an input $x$ is challenging to $D$. This \emph{variance score} is a macroscopic view of dropout variance in that it averages the uncertainty of $D$ wrt. $x$ over all classes.

\begin{definition}
The \textbf{variance score} of the input $x$ is given by
$$Var(x) = \frac{1}{C} \sum_{i=0}^{C}\operatorname{var}(P_{i}(x))$$
where $C$ is the number of classes and $\operatorname{var}$ denotes the standard variance function.
\end{definition}

A drawback of this metric is that it does not consider the prediction probabilities (thus, the actual distance to class boundaries). To overcome this, we propose a relative metric that normalizes the variance score with the highest probability output by $D$.

\begin{definition}
The \textbf{weighted variance score} of the input $x$ is
$$Var_w(x) =  \big ( \max_{i = 1:C} p_i (x) \big )^{-1} \cdot Var(x)$$
where $p_i (x)$ is the prediction probability of $x$ to class $i$ by $D$. 
\end{definition}

While variance and weighted variance scores of $x$ can be regarded as quantitative measures of the uncertainty of the model wrt. $x$, we also propose a nominal alternative. Instead of the variance of prediction probabilities, we focus on the actual class predictions produced by the different mutant models, that is, the classes with the highest probability scores. We construct a normalized histogram of these $k$ class predictions and we compare their distribution with that of a theoretical, worst-case, completely uncertain model, where the class predictions are uniformly distributed over all classes. Thus, in this worst case, the number of mutants predicting that an input $x$ belongs to class $i$ is approximately given by $\frac{k}{C}$.

To compare the actual class prediction distribution with the worst-case distribution, we rely on the discrete version of Kullback-Leibler (KL) divergence. When the uncertainty of $D$ is high (i.e. the mutants often disagree), the KL divergence is low.

\begin{definition}
The \textbf{Kullback-Leibler score} of the input $x$ is
$$KL(x) = \sum_{i=1}^{C}H_i \ln \frac{H_i}{Q_i}$$
where $i$ is the class label, $H$ is the normalized histogram, or frequencies, of the class predictions for $x$ given by the $k$ dropout mutants $M_1$ to $M_k$ of the original model $D$ and $Q$ is the uniform distribution (i.e. $Q_i = \frac{1}{C}$).
\end{definition}



\section{Experimental Setup}


\subsection{Objectives and Methodology}

\subsubsection{Test Selection}


Our first step is to assess the adequacy of each metric to select test inputs that challenge a given DL model $D$. To achieve this, we visualize the relation between the model accuracy when tested against the test inputs and the uncertainty of these inputs (we sort uncertainty in descending order). Observing this relation gives us a coarse-grained view of the trend between the two variables. More precisely, after sorting the test input set, we divide it into 10 equally-sized subsets. We compute the accuracy of the model wrt. the first subset, then iteratively add the subsequent subsets (thereby obtaining a higher proportion of the test data) and report each time the resulting accuracy values. For LSA, DSA, variance and weighted variance, a higher score means more uncertainty. For KL and probability, a lower score means more uncertainty. An adequate metric should sort data such that the accuracy increases at each iteration, as misclassifications should occur less frequently over the iterations. 

To corroborate our observations and make a fine-grained analysis of our data, we directly assess the existence of a statistical correlation between the metrics and misclassification. To do so, we encode the `correctness' of the prediction of $D$ for one particular input $x$ as a binary variable $b_x$ (well- or miss-classified). For each metric, we compute the correlation between the score obtained by all test inputs wrt. the metric and their corresponding binary variables. We use three types of correlation: Kendall, distance and Pearson correlations.

\subsubsection{Test Selection with Adversarial Data}

While the setup described above considers only real (original) data, we ask the question whether the introduction of artificial (adversarial) data has an impact on the capability of the metrics to pick up challenging test inputs. Adversarial data result from the successive application of minor perturbations to original data with the aim of deceiving a classifier. Those data have been a major concern~\cite{KurakinGB16a} and our metrics should be robust against them. Moreover, previous research \cite{ma2018deepgauge, Kim2019} also used adversarial data. To craft adversarial data, we use three established adversarial data generation algorithms: Fast Gradient Sign Method (FGSM)~\cite{Goodfellow2015}, Jacobian-based Saliency Map Attack (JSMA)~\cite{Papernot2016} and DeepFool (DF)~\cite{Moosavi-Dezfooli2016}. We apply each algorithm separately, each time doubling the size of the test input set (with as many adversarial data as original data) and use the aforementioned procedure to compute the Kendall, distance and Pearson correlations between misclassification and the metrics.

To demonstrate more finely the sensitivity of the metrics on adversarial data, we apply FGSM on 10 randomly selected images. Remember that this algorithm iteratively generates adversarial images (by mutating them - introducing noise), until it manages to change the outcome of the model. For each metric, we visualize the variation of the metric of the mutated images produced by the algorithm over the iterations in order to observe the trends. A monotonic evolution of the metric over the iterations would suggest the capability of this metric to discriminate the adversarial images ultimately produced from the original images and the ones produced at intermediate iterations. 

To confirm this hypothesis, we also study the existence of a statistical correlation between the metrics and misclassification wrt. adversarial inputs. We randomly select 100 test images and apply FGSM on each image. For each run, we retrieve the adversarial image obtained at the last iteration (which is misclassified) and the intermediate adversarial image obtained at the last but one iteration (which is still well classified). We associate each of the 200 resulting adversarial images with a binary variable (well- or mis-classified) and compute the Kendall, distance and Pearson correlations between the binary variables and the metrics of the corresponding images.

\subsubsection{Training Data Selection}

%

Having studied the adequacy of the metrics to select test inputs, we focus next on how much the selected test inputs can help to improve the model through retraining. To do this, we set up an iterative retraining process. 
At first, we randomly select a set of 10,000 images, which are to be used for the initial training. At each iteration, we add (without replacement) a batch of $5,000$ new images to the training set and retrain the model. The selected images are those that have the highest uncertainty (i.e. lowest score for $KL$ and $MaxP$, highest score for $Var$ and $Var_w$) or surprise (i.e. highest $LSA$ or $DSA$). We retrain the model based on the augmented training set for 200 epochs and compute the resulting accuracy wrt. the test data. We always ensure that test data are never used during any training nor any retraining. We repeat the process until there is no remaining data. 

To assess the adequacy of each metric, we observe the evolution of the accuracy wrt. the independent test data over the retraining iterations. Better metrics should yield higher and faster increases in accuracy. It is important to compare the accuracy achieved by the metrics trained on the same number of data in order to avoid any bias related to the size of the training set.  

To account for random variations due to the selection of the retraining data, we repeat the experiment five times and report the median accuracy values.

\subsection{Datasets}

We consider two datasets of image recognition problems. The first dataset concerns \emph{CIFAR-10} \cite{CIFAR10}, a collection of images that is widely used for research on DL. It contains 60,000 labelled images scattered in 10 classes, including 50,000 for training and 10,000 for testing. The DL model we use is a 10-layer \emph{VGG16} \cite{simonyan2014very} obtained by removing the top layers and inserting a batch normalization layer after each convolutional layer. We initialize the model with the pre-trained weights of VGG16 for ImageNet. We adopt stochastic gradient descent and a momentum of $0.9$ to optimize the weights. We also use learning rate decay, with an initial learning rate of $0.01$. We train the model for 300 epochs and keep the model with the best validation accuracy ($0.9232$).

The second dataset is \emph{Fashion-MNIST} \cite{MNIST}, a state-of-the-art clothes classification problem. It consists of 70,000 clothes images classified into 10 classes, separated in a training set of 60,000 images and a test set of 10,000 images. To perform the classification, we use \emph{Model 3} from DeepXplore \cite{Pei2017}, trained using Adam optimization method~\cite{Kingma2014} with typical parameter settings: an initial learning rate of $0.001$, the decay rates for the first and second moment estimates set to 0.9 and 0.999, and $\epsilon$ value set to 10E-8. We train Model 3 for 200 epochs and retain the model with the best validation accuracy $0.9288$. Note that the well-known MNIST handwriting dataset was discarded because we deem it too simple for our purpose, as we could easily get models with $100\%$ validation accuracy. Fashion-MNIST was therefore used as a more challenging replacement. Henceforth, we denote Fashion-MNIST as MNIST for simplicity.

\subsection{Implementation}

We tooled our approach on top of Keras and Tensorflow, and use the third library Foolbox \cite{DBLP:journals/corr/RauberBB17} to generate adversarial images. All experiments were run on GPU K80. 

When the considered DL model (e.g. VGG16) does not use standard dropout for training, we implement permanent dropout as Lambda layers. When the DL model includes Dropout layers for training (e.g. Model 3), we simply keep those Dropout layers working during testing. 

In whichever case, the only hyperparameters of our method are the dropout rate $r$ (probability of dropping-out neurons) and the number $k$ of mutant models. If $r$ is too large, the original model competence will be significantly degraded, which will result in poor quality mutants. On the contrary, a small $r$ results in too small variations for our method to perform well. For the clipped VGG16, we experimentally set the drop rate to 0.2. As for Model 3, we keep the same rate as the one it uses by default for training.

The dropout layers make the model generate different outputs each time a given input $x$ is tested, by randomly masking the neurons on the fly. Thereby, it simulates the production of mutant models and the passing of $x$ into each of these mutants. The number of mutants thus corresponds to the number of times we make $x$ go through the model with dropout layers enabled. 

According to the law of the large numbers, the variance of a sample better approximates the theoretical variance with more repetitions. In our experiments, $k = 50$ appeared as a good trade-off between convergence of the variance and computation cost.

We implemented the surprise adequacy metrics ($LSA$ and $DSA$) using the established libraries \textsc{NumPy} and \textsc{scikit-learn}.

\subsection{Threats to Validity}

\subsubsection{Internal}

The main threats of internal validity concern the implementation of the software constituents of our study. 

Some are addressed by the fact that we reuse existing model architectures with typical parameterizations. The resulting models obtain a high accuracy on state-of-the-art datasets used as is (including their splitting into training and test sets), which indicates that our setup was appropriate. 

We implemented dropout ``from scratch'' (i.e. as Lambda layers) in one case and, in the other case, we reused the implementation natively embedded in the training process. The use of these two alternatives increases our confidence in the validity of our results.

Finally, the implementation of the different metrics was tested manually and through various experiments. Regarding LSA, it has been shown that the choice of the layers has an impact on the adequacy of the metric~\cite{Kim2019}. However, Kim et al. could find no correlation with the depth of the layer. As such, we make the same choice as Kim et al. and compute LSA on the deepest hidden layer.

We evaluate our metrics using test sets of real data, adversarial data, and a mix of real and adversarial data in an attempt to make a thorough and unbiased evaluation. Also, during the analysis of the adversarial images, we selected only 10 images on which we applied the adversarial generation algorithms in order to reduce computation cost. Still, the results we obtained are in line with the correlations, which were computed with 200 adversarial inputs (also randomly selected). 

The retraining process also involves randomness. However, we  repeated it five times and observed no outlier case.


\subsubsection{External}

The threats to external validity originate from the number of datasets, models and adversarial generation algorithms we considered. 
The settings we used are established in the scientific literature and allow the comparison of our approach with the related work. The replication and the complementation of our study are further facilitated by black-box nature of our metrics: all of them necessitate only the prediction probabilities to compute standard statistical measures. 




\subsubsection{Construct}
 Construct validity threats originate from the measurements we consider. We consider the correlation between the studied metrics and misclassification, which is a natural metric to use (in some sense equivalent to the fault detection and test criteria relations studied by software engineering literature \cite{GligoricGZSAM13, AndrewsBL05}). We also visualize our data wrt. classification accuracy to demonstrate our results and the observed trends. Moreover, we compare with surprise adequacy \cite{Kim2019}, which is the current state-of-the-art method. We do not include coverage criteria like \cite{Pei2017,ma2018deepgauge} in our comparison, because those are not directed towards test selection.

\section{Results}

\subsection{Test Selection with Real Data}

\begin{figure*}[!ht]
  \centering
  \vspace{-1.0em}
\subfloat[MNIST]{ \includegraphics[width=0.85\linewidth]{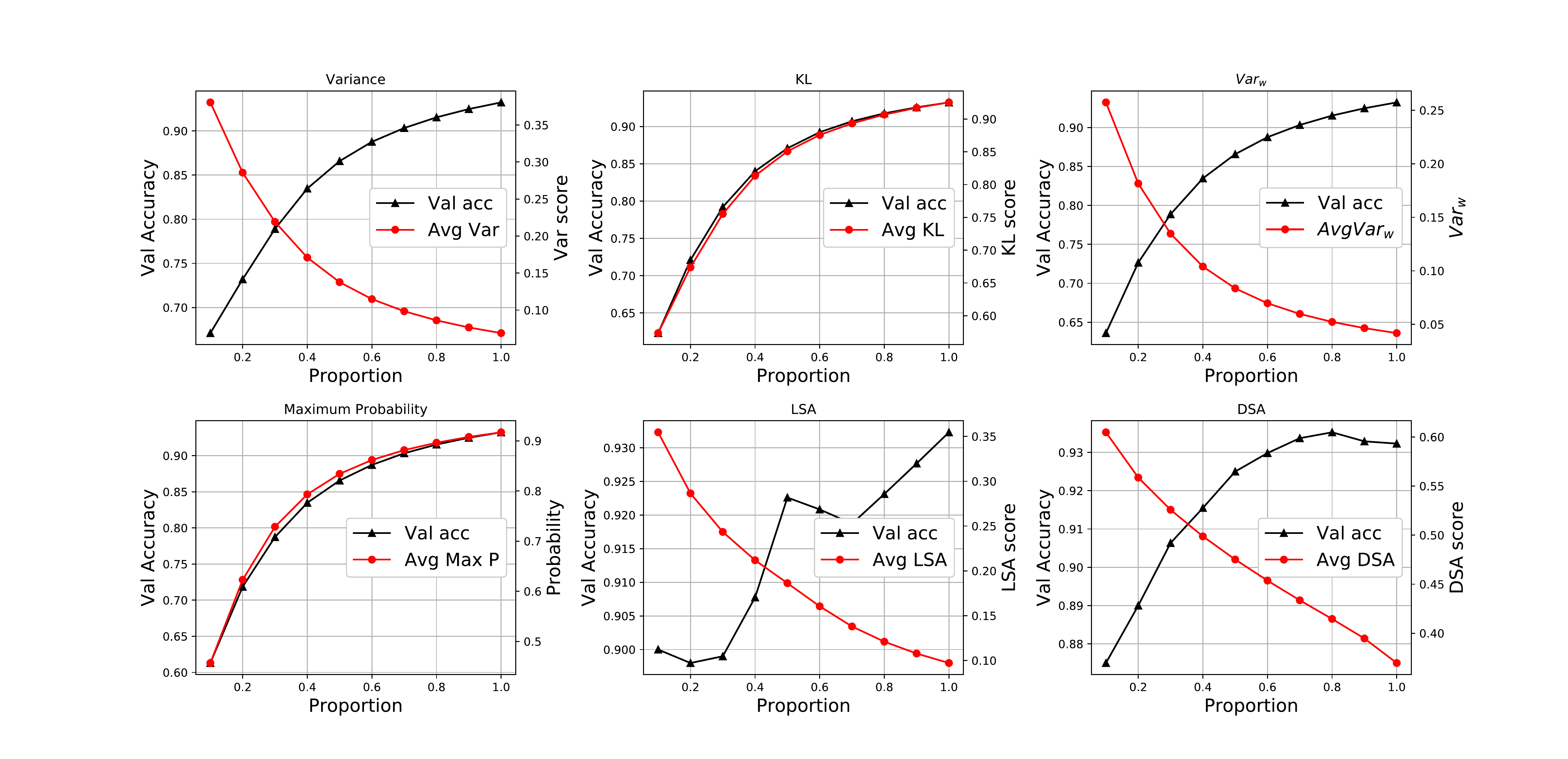}}\\
\vspace{-1.0em}
	\subfloat[CIFAR-10]{ \includegraphics[width=0.85\linewidth]{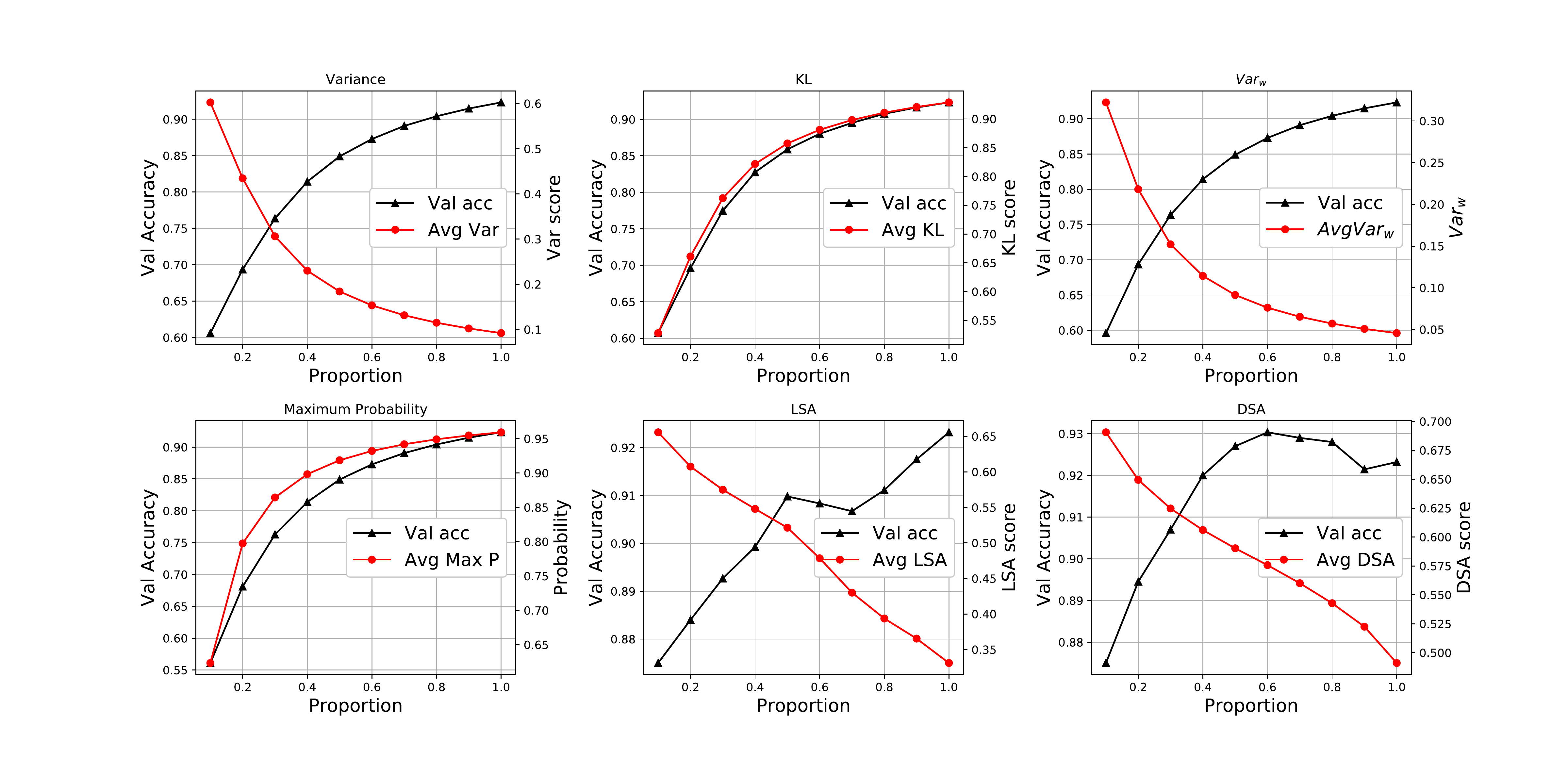}}
	\caption{Validation accuracy (Y-axis, left) obtained as the model is tested against a greater proportion of test data (X-axis) sorted according to the metrics (Y-axis, right). Black lines represent the classification accuracy curves, while red lines represent the metric curves. For all metrics except LSA and DSA, the accuracy increases in a monotonic way, indicating that the four uncertainty-based metrics appropriately select more challenging test data.}
	\Description{RQ1 trends}
	\label{fig:RQ1_acc_score}
\end{figure*}

\subsubsection{Visualization}

Figure \ref{fig:RQ1_acc_score} demonstrates the relation between classification accuracy and the studied metrics for the MNIST and CIFAR-10 benchmarks. Each graph corresponds to a metric and shows how accuracy increases as a greater proportion of the original test data (sorted by descending uncertainty / surprise according to the metrics) is used (from 10\% to 100\%, with steps of 10\%). The X-axis represents the proportion of test data used. The Y-axis represents the classification accuracy (left side of the graphs), denoted as `Val acc' and marked in black, and the average of the metric (right side of the graphs), marked in red. 


From these figures we can observe that all $Var$, $KL$, $Var_w$ and $MaxP$ have similar trends: when the average uncertainty (represented by the metrics) decreases, the accuracy increases at a similar rate. This observation confirms our intuition that the selected techniques represent the relative challenge introduced by the selected test data. In other words, ordering the test data according to the uncertainty they engender can prioritize the most challenging ones, i.e., the most likely to make the classifier do a mistake.  

When it comes to $LSA$ and $DSA$, however, the trends are not confirmed: while the accuracy improves overall, it drops at intermediate stages (LSA) or from the second half (DSA). This indicates that, in such settings, the improvement allowed by these surprise-based metrics are incidental.

\begin{table}[!ht]
	\caption{Correlation between misclassification and metrics on the real test data (CIFAR-10 \& MNIST).}
	\label{tab:cor}
	\begin{tabular}{ l l l l l l l l }
		\hline
		\scriptsize{Dataset}    & \scriptsize{Correlation} & \scriptsize{KL}  & \scriptsize{Var}  & \scriptsize{LSA}   & \scriptsize{DSA}     & \scriptsize{$Var_w$}  & \scriptsize{P}   \\ \hline	
		\multirow{3}{*}{\scriptsize{CIFAR-10}} 
		& \scriptsize{Kendall}     & \scriptsize{0.4158} & \scriptsize{-0.3024} & \scriptsize{-0.0778} & \scriptsize{-0.0064} & \scriptsize{-0.3743} & \scriptsize{0.3205} \\ 
		& \scriptsize{Distance}    & \scriptsize{0.4642} & \scriptsize{0.4607}  & \scriptsize{0.0620}   & \scriptsize{0.0494}  & \scriptsize{0.4758}  & \scriptsize{0.4650}  \\ 
		& \scriptsize{Pearson}     & \scriptsize{0.4685} & \scriptsize{-0.4632} & \scriptsize{-0.0674} & \scriptsize{-0.0171} & \scriptsize{-0.4797} & \scriptsize{0.4444} \\ \hline
			
	\multirow{3}{*}{\scriptsize{MNIST}} 
	    & \scriptsize{Kendall}      & \scriptsize{0.3836} & \scriptsize{-0.2845} & \scriptsize{-0.069}  & \scriptsize{-0.0340}  & \scriptsize{-0.3179} & \scriptsize{0.3077} \\ 
		& \scriptsize{Distance}    & \scriptsize{0.4598} & \scriptsize{0.4308}  & \scriptsize{0.0694}  & \scriptsize{0.0647}  & \scriptsize{0.4586}  & \scriptsize{0.4787} \\ 
		& \scriptsize{Pearson}     & \scriptsize{0.4772} & \scriptsize{-0.4251} & \scriptsize{-0.0567} & \scriptsize{-0.0504} & \scriptsize{-0.4611} & \scriptsize{0.4879} \\ \hline
		
	\end{tabular}
\end{table}

\subsubsection{Correlations} 

Table \ref{tab:cor} records the Kendall, distance and Pearson correlations between the metrics and misclassification. We observe that KL, $Var$, $Var_w$ and $MaxP$ have a medium degree of correlation, meaning that they can lead to valuable test data, i.e., those causing misclassifications. Conversely, we observe that both LSA and DSA have very low correlation to misclassification. This can be explained by the fact that LSA and DSA make a diverse test selection over the feature space that is not necessarily close to the boundaries between the classes. 

Overall, these results confirm our previous observations: KL, $Var$, $Var_w$  and $P$ correlate with misclassifications. However, this correlation is only moderate, meaning that it cannot perfectly distinguish between well-classified and misclassified inputs.

\subsection{Test Selection with Adversarial Data}
 
\begin{table}[ht!]
	\caption{Correlation between misclassification and metrics on a test set comprising real and adversarial data (CIFAR-10 \& MNIST - 10,000 real data and 5,000 adversarial.}
	\label{tab:cor_adversarial}
	\begin{tabular}{ l l l l l l l l }
		\hline
		\scriptsize{Dataset}    & \scriptsize{Correlation} & \scriptsize{KL}  & \scriptsize{Var}  & \scriptsize{LSA}   & \scriptsize{DSA}     & \scriptsize{$Var_w$}  & \scriptsize{P}   \\ \hline	
		\multicolumn{1}{c}{\multirow{3}{*}{\begin{tabular}[c]{@{}c@{}}\scriptsize{CIFAR-10 + DF}\end{tabular}}} & \scriptsize{Kendall}  & \scriptsize{0.577}  & \scriptsize{-0.5505} & \scriptsize{-0.1029} & \scriptsize{-0.0196} & \scriptsize{-0.6251} & \scriptsize{0.6587} \\ 
		\multicolumn{1}{c}{}                                                                         & \scriptsize{Distance}    & \scriptsize{0.6088} & \scriptsize{0.6200}    & \scriptsize{0.1695}  & \scriptsize{0.0664}  & \scriptsize{0.6709}  & \scriptsize{0.8630}  \\ 
		\multicolumn{1}{c}{}                                                                         & \scriptsize{Pearson}     & \scriptsize{0.5741} & \scriptsize{-0.6074} & \scriptsize{-0.0968} & \scriptsize{-0.0285} & \scriptsize{-0.6469} & \scriptsize{0.8622} \\ \hline
		\multirow{3}{*}{\begin{tabular}[c]{@{}l@{}}\scriptsize{CIFAR-10 + FGSM}\end{tabular}}                     & \scriptsize{Kendall}  & \scriptsize{0.6705} & \scriptsize{-0.6052} & \scriptsize{-0.1358} & \scriptsize{-0.0138} & \scriptsize{-0.6861} & \scriptsize{0.6639} \\ 
		& \scriptsize{Distance}    & \scriptsize{0.7361} & \scriptsize{0.7496}  & \scriptsize{0.1824}  & \scriptsize{0.0867}  & \scriptsize{0.7856}  & \scriptsize{0.8735} \\ 
		& \scriptsize{Pearson}    & \scriptsize{0.7027} & \scriptsize{-0.7371} & \scriptsize{-0.1423} & \scriptsize{-0.0160}  & \scriptsize{-0.7505} & \scriptsize{0.8683} \\ \hline
		\multirow{3}{*}{\begin{tabular}[c]{@{}l@{}}\scriptsize{CIFAR-10 + JSMA}\end{tabular}}                     & \scriptsize{Kendall}  & \scriptsize{0.6260}  & \scriptsize{-0.5823} & \scriptsize{-0.1305} & \scriptsize{0.0045}  & \scriptsize{-0.6566} & \scriptsize{0.6637} \\ 
		& \scriptsize{Distance}    & \scriptsize{0.6746} & \scriptsize{0.688}   & \scriptsize{0.1855}  & \scriptsize{0.0675}  & \scriptsize{0.7282}  & \scriptsize{0.8696} \\ 
		& \scriptsize{Pearson}     & \scriptsize{0.643}  & \scriptsize{-0.6771} & \scriptsize{-0.1311} & \scriptsize{-0.0012} & \scriptsize{-0.7034} & \scriptsize{0.8650}  \\ \hline

\multicolumn{1}{c}{\multirow{3}{*}{\begin{tabular}[c]{@{}c@{}}\scriptsize{MNIST + DF}\end{tabular}}} & \scriptsize{Kendall}  & \scriptsize{0.5028} & \scriptsize{-0.4246} & \scriptsize{-0.0736} & \scriptsize{-0.007}  & \scriptsize{-0.4529} & \scriptsize{0.4340}  \\ 
		\multicolumn{1}{c}{}                                                                         & \scriptsize{Distance}    & \scriptsize{0.5779} & \scriptsize{0.5449} & \scriptsize{0.0608}  & \scriptsize{0.0578}  & \scriptsize{0.5431}  & \scriptsize{0.5669} \\ 
		\multicolumn{1}{c}{}                                                                         & \scriptsize{Pearson}     & \scriptsize{0.5650}  & \scriptsize{-0.5130}  & \scriptsize{-0.0414} & \scriptsize{-0.0169} & \scriptsize{-0.4638} & \scriptsize{0.5505} \\ \hline

		\multirow{3}{*}{\begin{tabular}[c]{@{}l@{}} \scriptsize{MNIST + FGSM} \end{tabular}}                  & \scriptsize{Kendall}  & \scriptsize{0.7088} & \scriptsize{-0.6134} & \scriptsize{-0.1286} & \scriptsize{-0.1093} & \scriptsize{-0.6686} & \scriptsize{0.6695} \\ 
		& \scriptsize{Distance}    & \scriptsize{0.8247} & \scriptsize{0.7747}  & \scriptsize{0.1125}  & \scriptsize{0.1475}  & \scriptsize{0.7881}  & \scriptsize{0.883}  \\ 
		& \scriptsize{Pearson}     & \scriptsize{0.8054} & \scriptsize{-0.7360}  & \scriptsize{-0.089}  & \scriptsize{-0.1338} & \scriptsize{-0.6991} & \scriptsize{0.8766} \\ \hline

		\multirow{3}{*}{\begin{tabular}[c]{@{}l@{}}\scriptsize{MNIST + JSMA}\end{tabular}}                     & \scriptsize{Kendall}  & \scriptsize{0.6720}  & \scriptsize{-0.6056} & \scriptsize{-0.0865} & \scriptsize{-0.0863} & \scriptsize{-0.6647} & \scriptsize{0.6701} \\ 
		& \scriptsize{Distance}    & \scriptsize{0.7816} & \scriptsize{0.7701}  & \scriptsize{0.0790}   & \scriptsize{0.1217}  & \scriptsize{0.8035}  & \scriptsize{0.8816} \\ 
		& \scriptsize{Pearson}     & \scriptsize{0.7563} & \scriptsize{-0.7421} & \scriptsize{-0.0381} & \scriptsize{-0.1081} & \scriptsize{-0.7153} & \scriptsize{0.8735} \\ \hline

	\end{tabular}
\end{table}

\subsubsection{Mix of Real and Adversarial Data}

 Table \ref{tab:cor_adversarial} records the correlations between the metrics and misclassification when the original test data set is augmented with adversarial data. Interestingly, the correlations of all metrics are stronger than they were with real data only. KL, $Var$, $Var_w$ and $MaxP$ now have a strong degree of correlation with misclassifications, while the correlation of LSA and DSA remain weak. This indicates that our uncertainty-based metrics can distinguish between real and adversarial data. We also infer that adversarial data do not form a challenging scenario to evaluate test selection methods (as performed by related work \cite{ma2018deepgauge, Pei2017, Kim2019}). 
 
The above result can be explained by the fact that adversarial images have some form of artificial noise that the classifier never experienced during training. This noise makes the classifier less confident on how to deal with them, a fact reflected by the metrics. Given that the adversarial images are misclassified, it is possible that the metrics are even more appropriate to distinguish between adversarial data and real data than they are to differentiate well- and miss-classified data. To further dig into this question, we investigate the relation between the metrics and adversarial data.

\subsubsection{Well- and Miss-Classified Adversarial Data}

\begin{figure*}[h]
	\centering
	\vspace{-1.0em}
	\subfloat[MNIST]{ 
	\begin{tabular}{c c c}
		 { \includegraphics[scale=0.28]{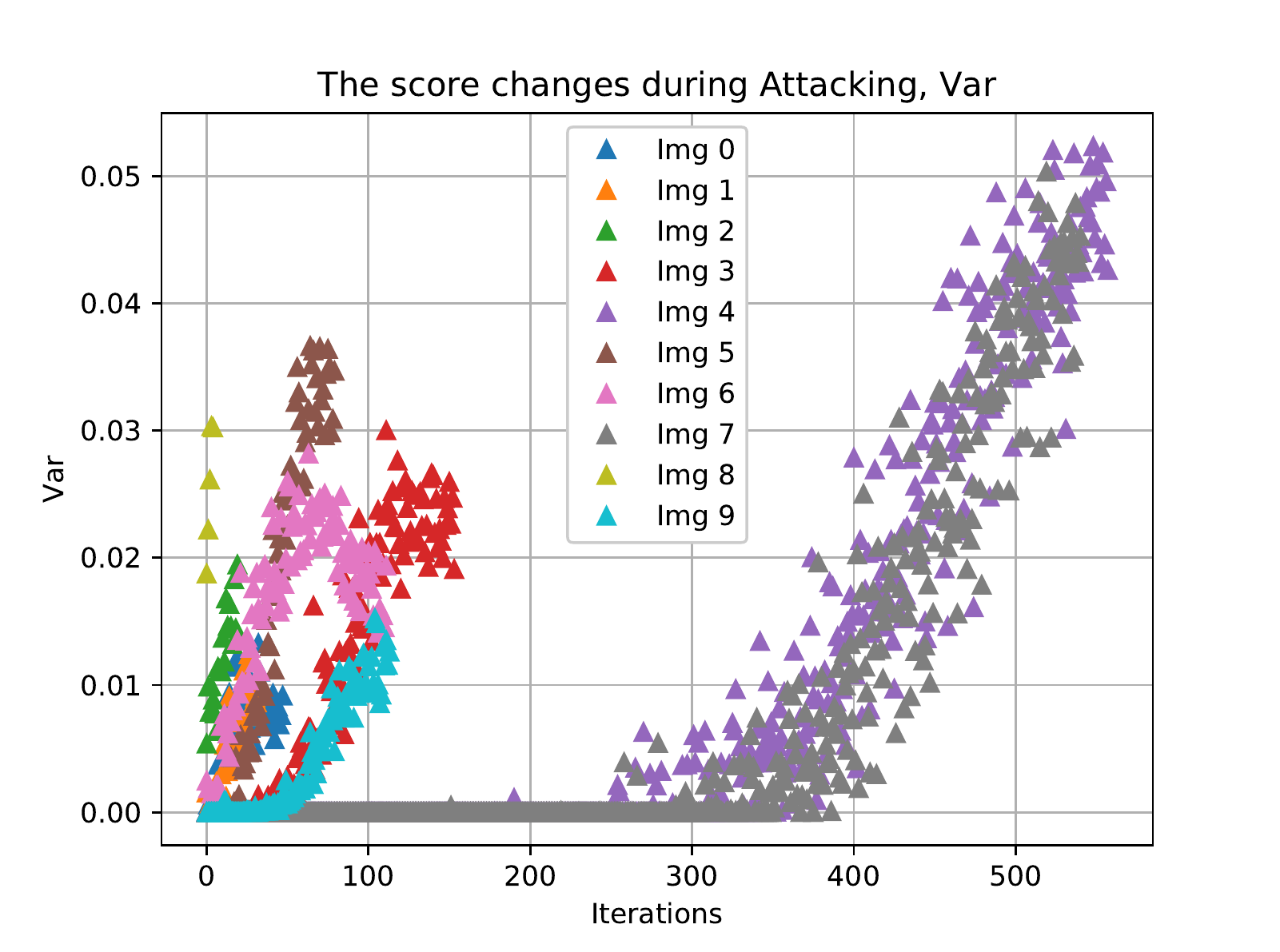} }
		&
	     { \includegraphics[scale=0.28]{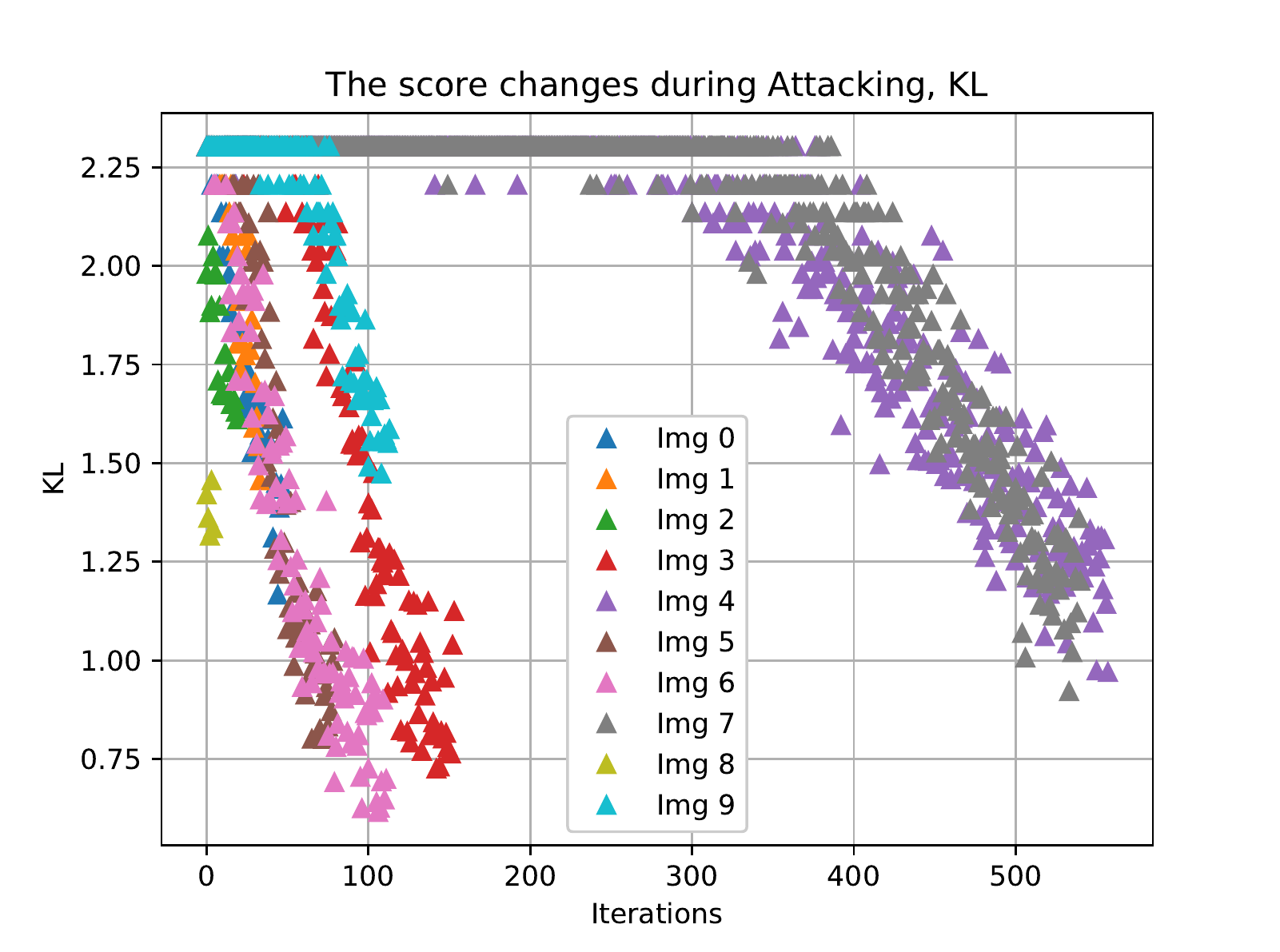} }
		&
		 { \includegraphics[scale=0.28]{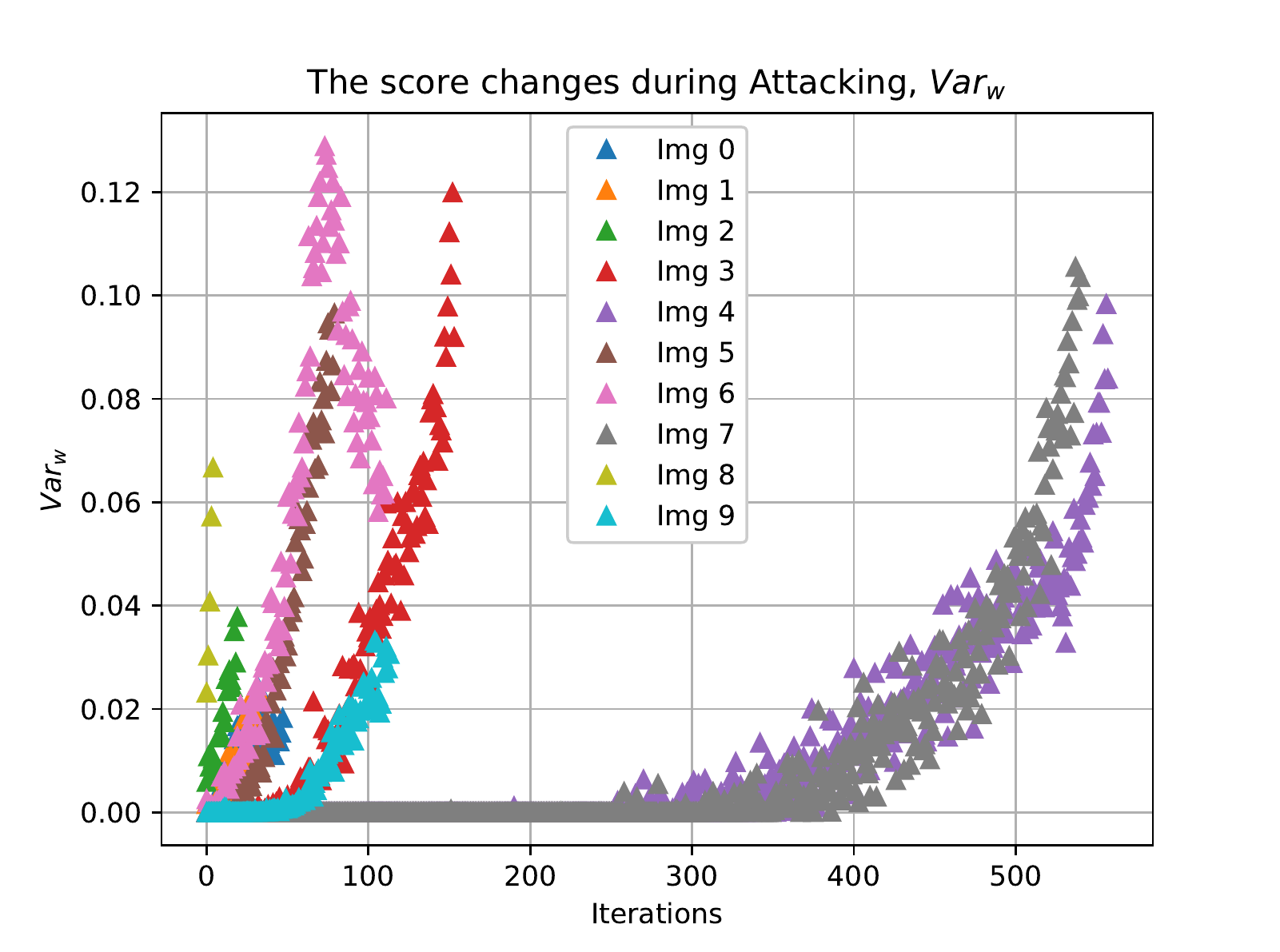} }
		\\
		 { \includegraphics[scale=0.28]{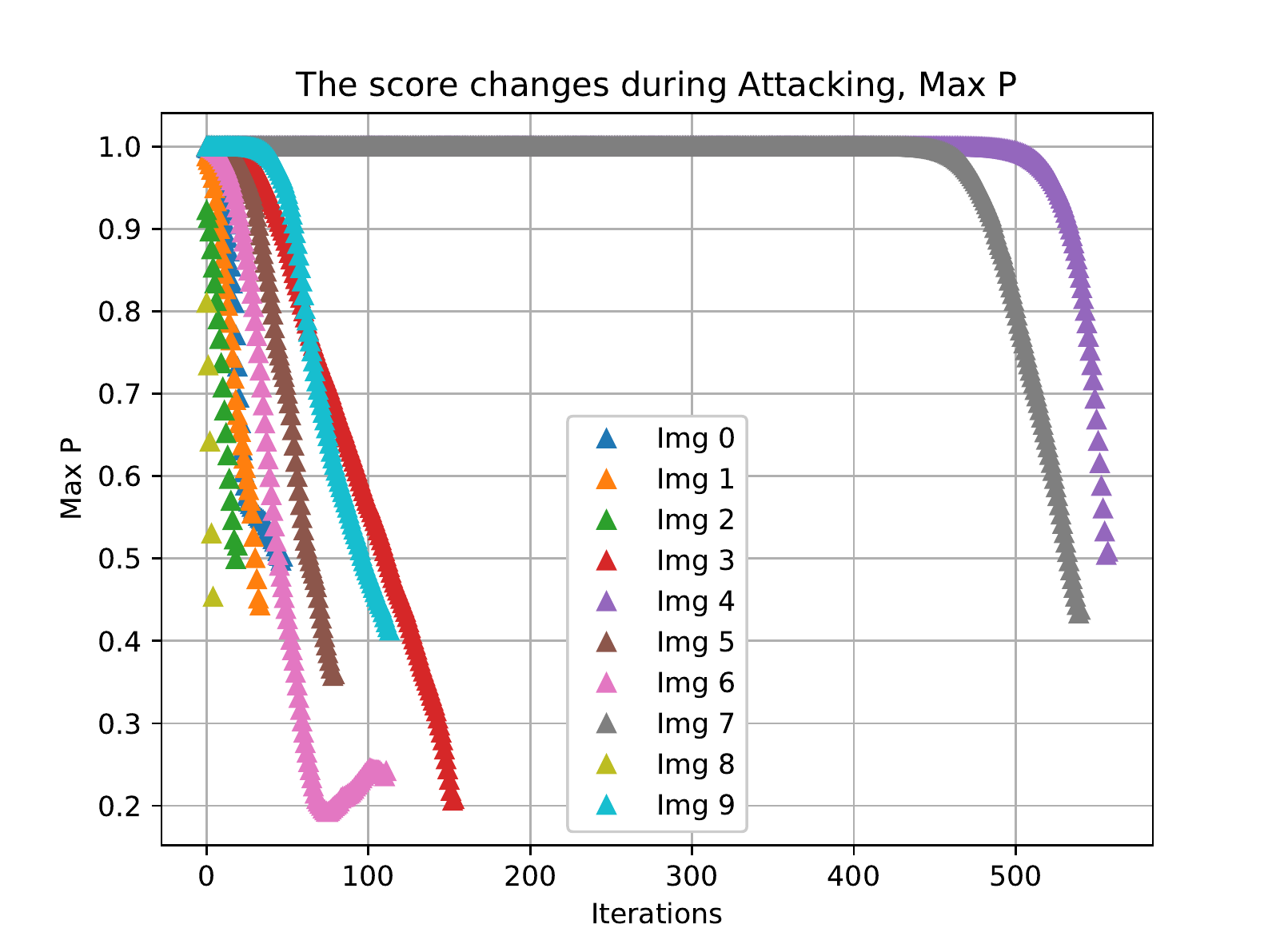} }
		&
		 { \includegraphics[scale=0.28]{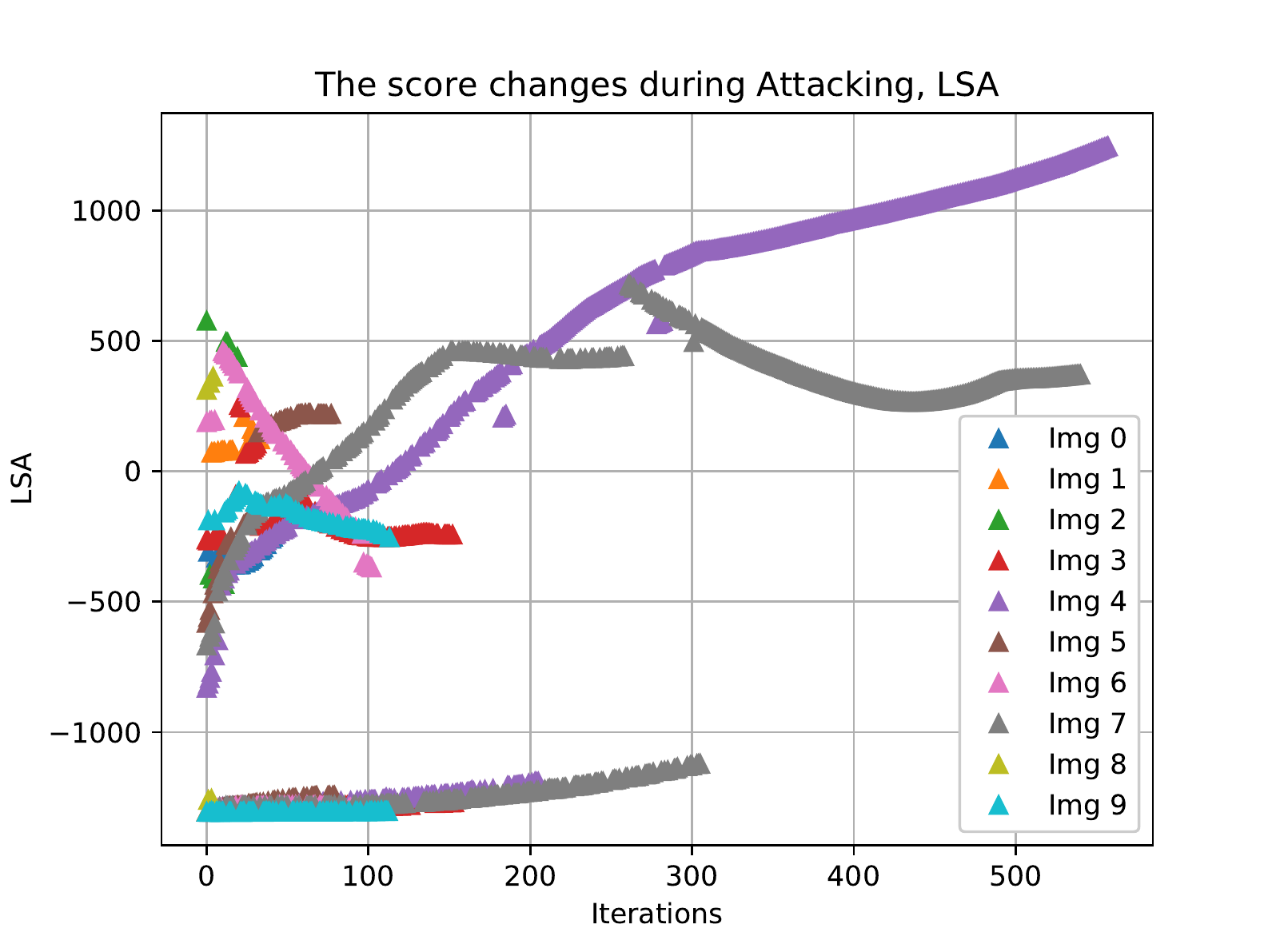} }
		&
		{ \includegraphics[scale=0.28]{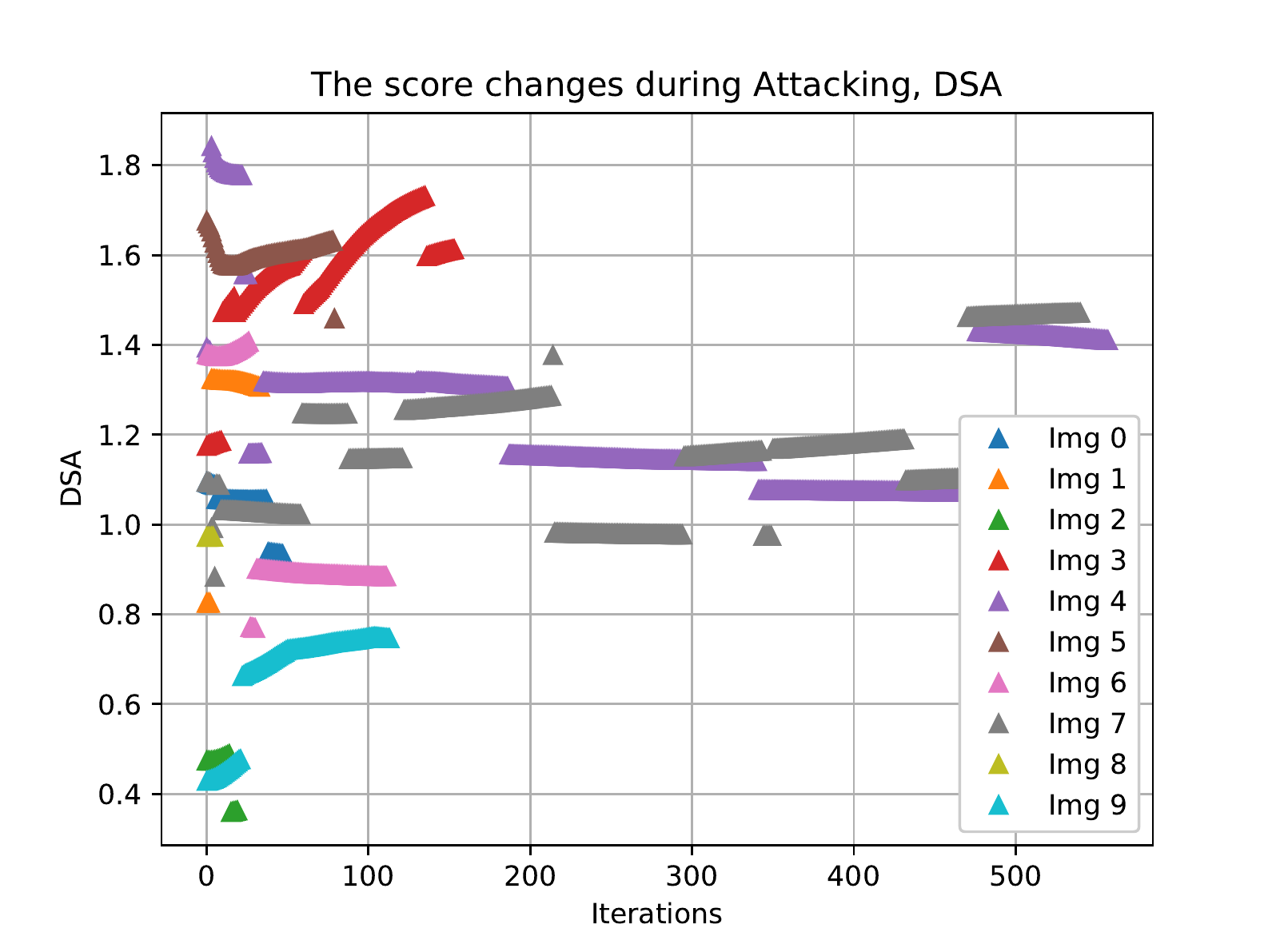} }
	\end{tabular}}\\
	\vspace{-1.0em}
		\subfloat[CIFAR-10]{ 
	\begin{tabular}{c c c}
		{ \includegraphics[scale=0.28]{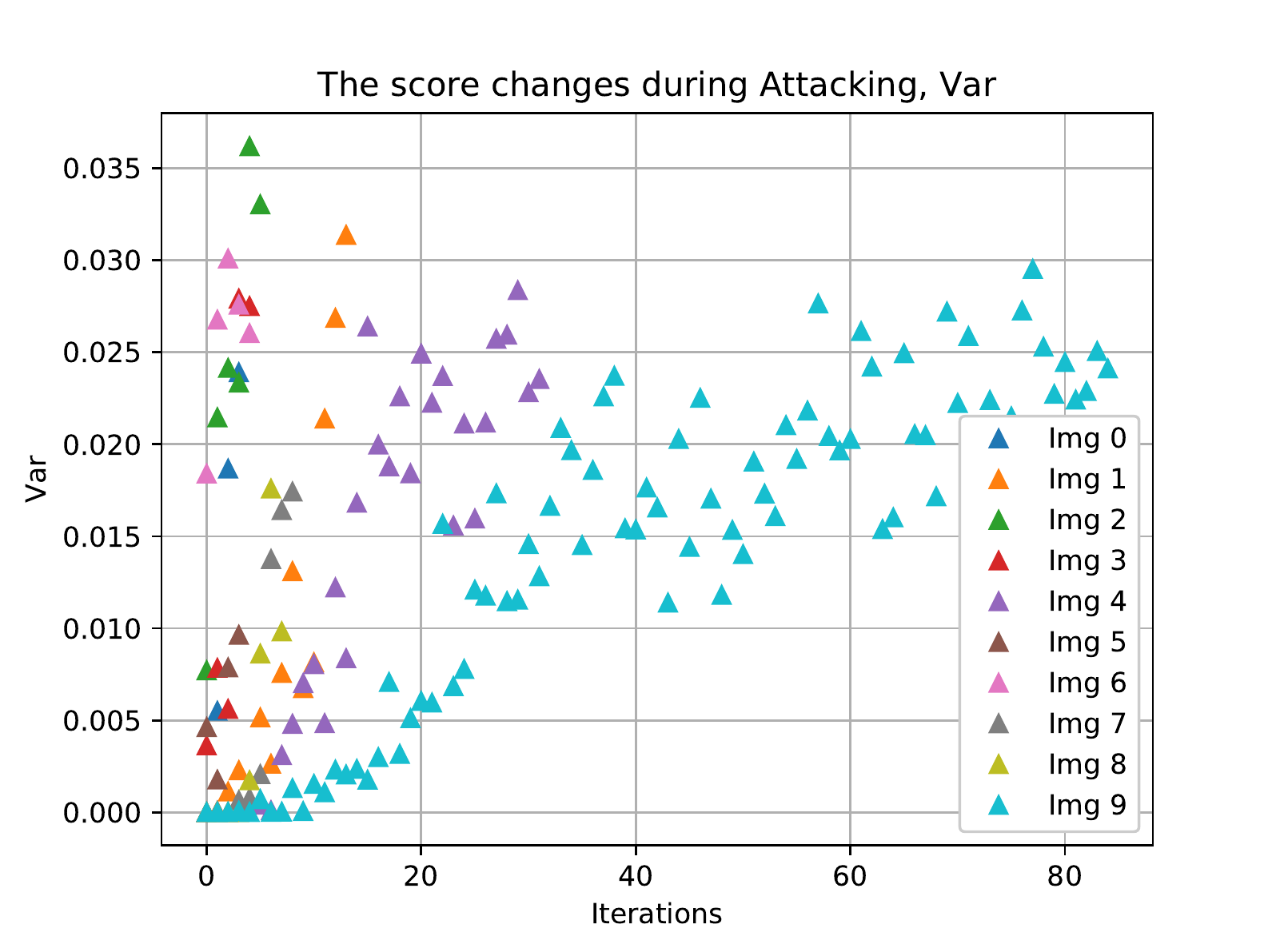} }
		&
		{ \includegraphics[scale=0.28]{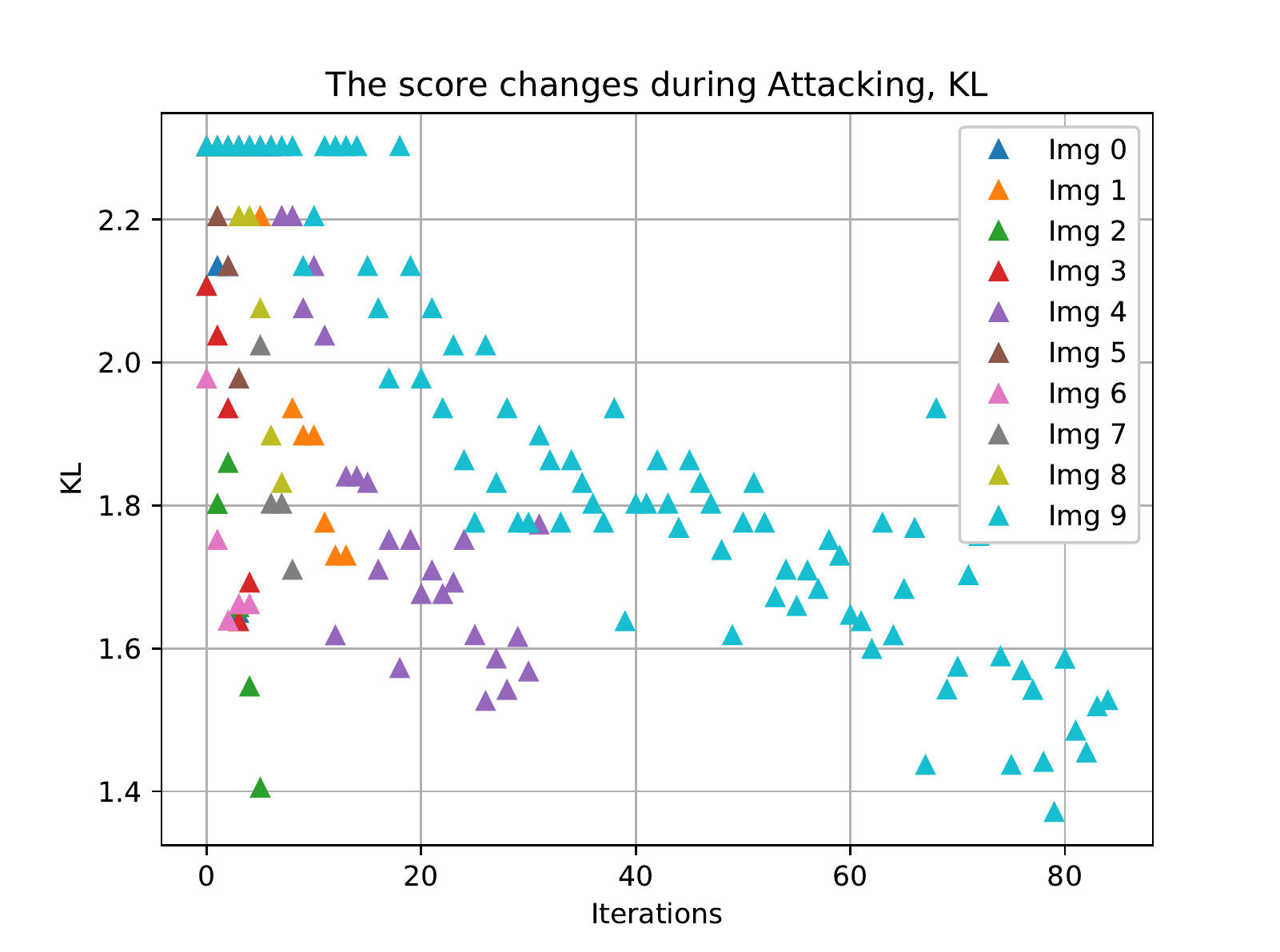} }
		&
		 { \includegraphics[scale=0.28]{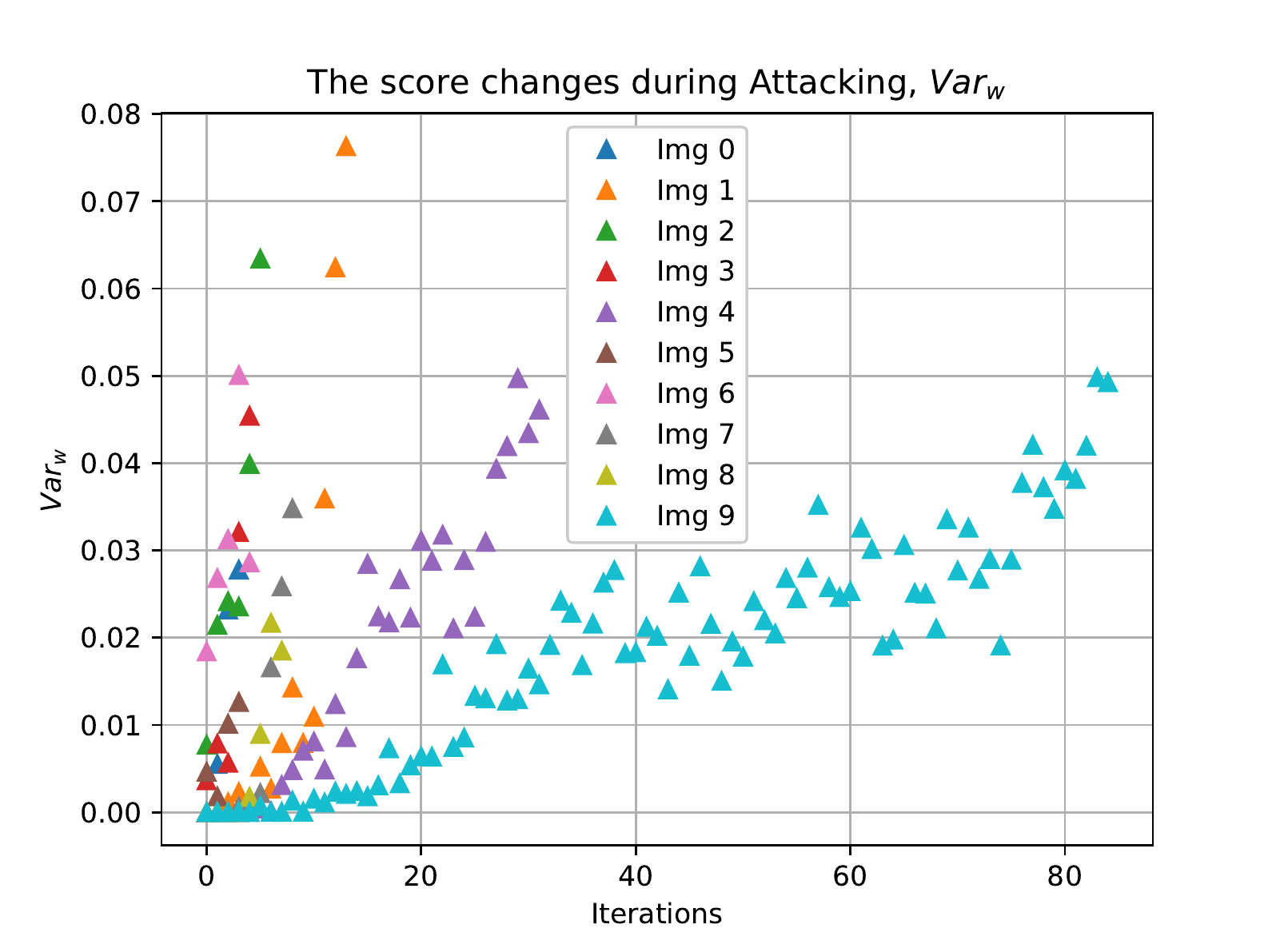} }
		\\
		 { \includegraphics[scale=0.28]{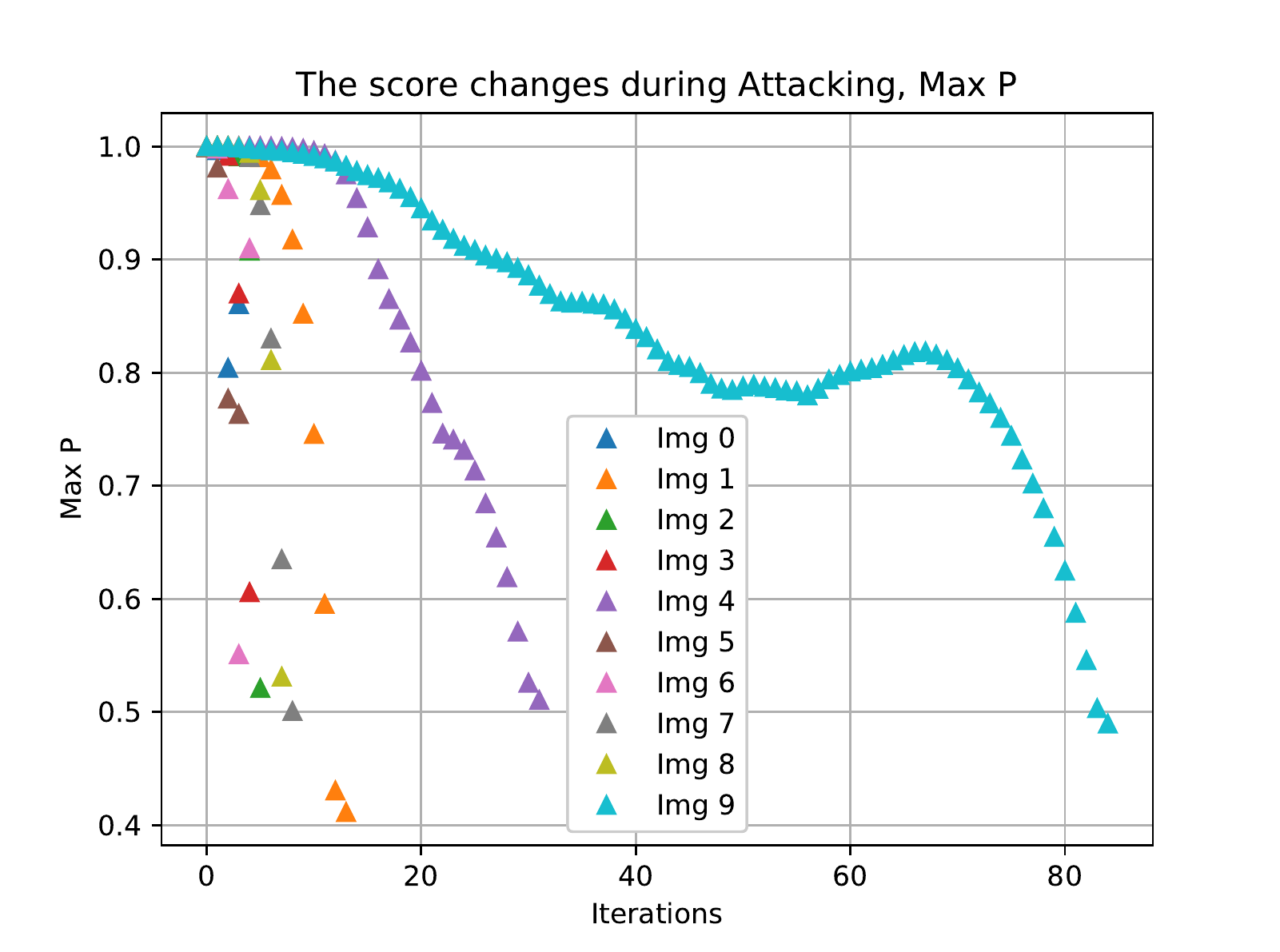} }
		&
		 { \includegraphics[scale=0.28]{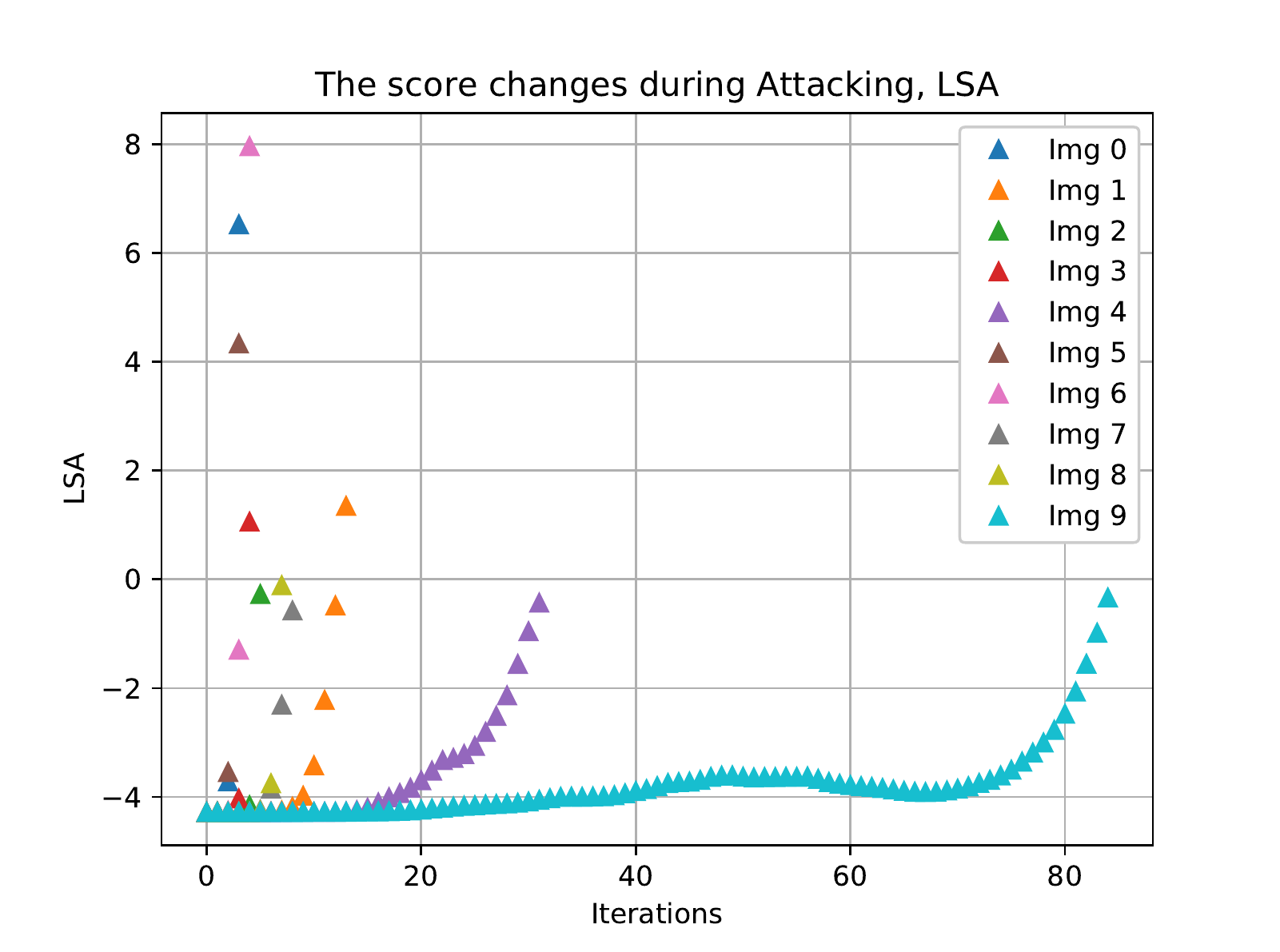} }
		&
		{ \includegraphics[scale=0.28]{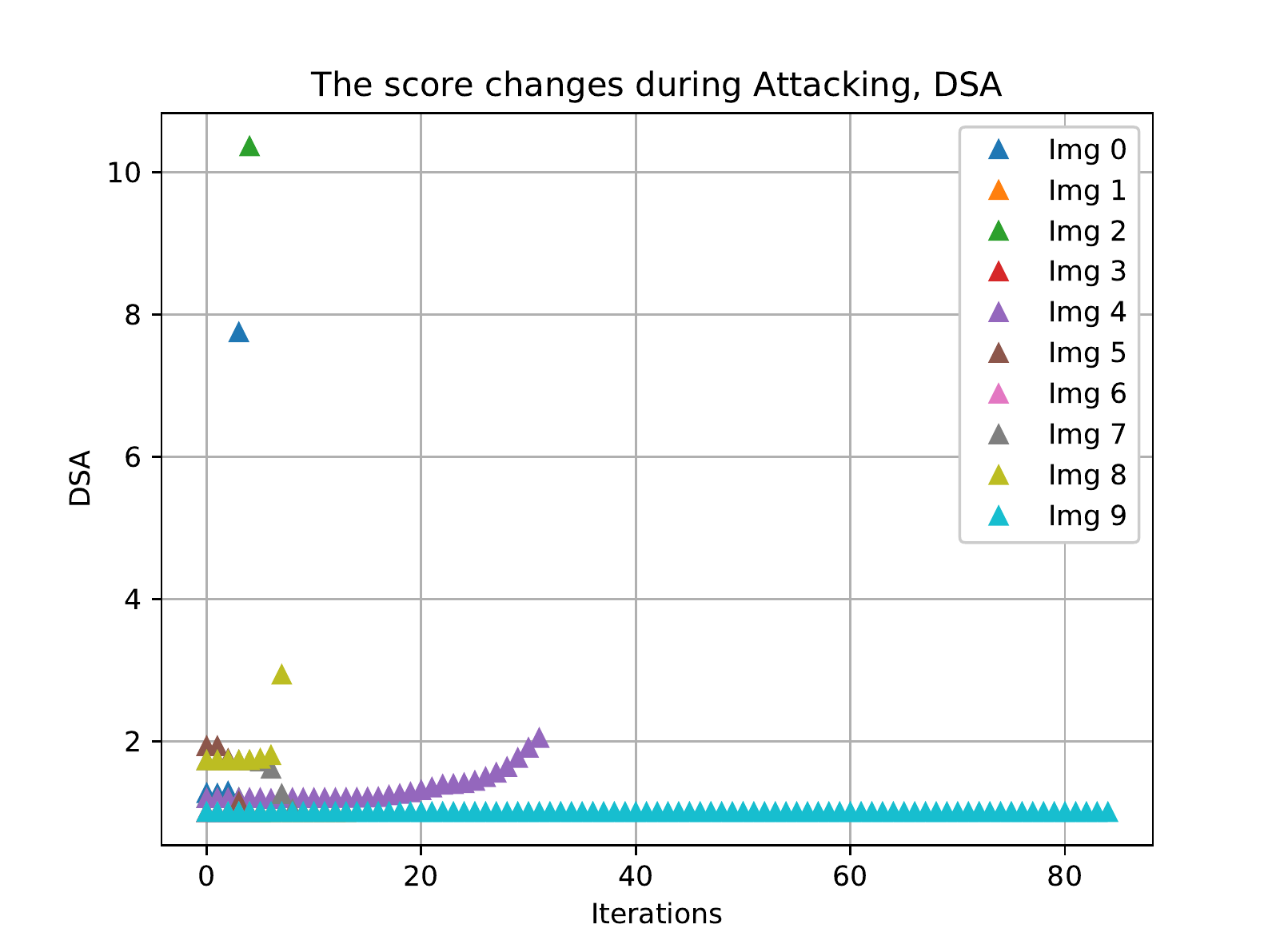} }
		
	\end{tabular}
	}
	\caption{Variations of each metric (Y-axis) over the iterations (X-axis) of the FGSM algorithm. All metrics except DSA and LSA capture well the adversarial data generation process.}
	\label{fig:score_adv_intermediate_images}
\end{figure*}

Figure \ref{fig:score_adv_intermediate_images} shows how each metric (Y-axis) varies over the iterations of the FGSM algorithm (X-axis) applied to 10 original images. We can see that, according to all uncertainty-based metrics, FGSM progressively increases the uncertainty of the images, to the point that in the latter iterations (i.e. when the image almost fails the classifier), the uncertainty has substantially increase wrt. the original image. Thus, those metrics capture well the adversarial generation process, which tends to indicate that they can indeed detect adversarial images. This does not seem to be the case for surprise-based metrics, though.

To pursue our investigations, we randomly selected 100 images from each of our two datasets and apply FGSM. In addition to each generated (misclassified) image, we keep one of the intermediary images generated by the algorithm (which is thus well-classified). Then, we use the resulting 200 artificial images to compute the correlation between our metrics and misclassification. 

Table \ref{AdvCorr} shows our results. 
Surprisingly, for both datasets none of the metrics perform well. This finding suggests that, when it comes to adversarial data only, the metrics have a hard time discriminating well-classified and miss-classified data due to their inherent noise (properties not encountered during training). 

Joining together all our previous findings, we conclude that our uncertainty-based metrics are able to distinguish between real, well-classified inputs and all (real and adversarial) misclassified inputs. It can achieve this with more ease as the misclassified inputs are artificial. When confronted to adversarial data only, however, they lose their capability. This is likely due to the fact that, as we observed in Figure \ref{fig:score_adv_intermediate_images}, in many cases the adversarial generation algorithm rapidly introduces a lot of uncertainty within the inputs that it continues to modify iteratively.



\begin{table}[]
	\caption{Correlation between metrics and misclassification using  100 well classified (adversarial) and 100 misclassified (adversarial) inputs.}
	\label{AdvCorr}
\vspace{-1.0em}
	\begin{tabular}{ l l l l l l l l }
		\hline
		\scriptsize{Dataset}    & \scriptsize{Correlation} & \scriptsize{KL}  & \scriptsize{Var}  & \scriptsize{LSA}   & \scriptsize{DSA}     & \scriptsize{$Var_w$}  & \scriptsize{P}   \\ \hline	
		
\multirow{3}{*}{\begin{tabular}[c]{@{}c@{}}\scriptsize{CIFAR-10} \end{tabular}} 
	&	\scriptsize{Kendall}  & \scriptsize{0.6419} & \scriptsize{-0.6508} & \scriptsize{0.1432} & \scriptsize{0.0498} & \scriptsize{-0.5888} & \scriptsize{0.1995} \\ 
	&	\scriptsize{Distance}    & \scriptsize{0.8099} & \scriptsize{0.7762}  & \scriptsize{0.1972} & \scriptsize{0.1175} & \scriptsize{0.6994}  & \scriptsize{0.2494} \\ 
	&	\scriptsize{Pearson}   & \scriptsize{0.7906} & \scriptsize{-0.7588} & \scriptsize{0.1622} & \scriptsize{0.0358} & \scriptsize{-0.6980}  & \scriptsize{0.1733} \\ \hline
	
	\multirow{3}{*}{\begin{tabular}[c]{@{}c@{}}\scriptsize{MNIST} \end{tabular}}
	 & \scriptsize{Kendall}  & \scriptsize{0.0324} & \scriptsize{-0.0048} & \scriptsize{-0.0235} & \scriptsize{0.0631} & \scriptsize{-0.0465} & \scriptsize{0.0325} \\ 
	&	\scriptsize{Distance}    & \scriptsize{0.065}  & \scriptsize{0.0554}  & \scriptsize{0.0715}  & \scriptsize{0.0912} & \scriptsize{0.0880}   & \scriptsize{0.1604} \\ 
	&	\scriptsize{Pearson}     & \scriptsize{0.0335} & \scriptsize{-0.0076} & \scriptsize{-0.0537} & \scriptsize{0.0700}   & \scriptsize{-0.0724} & \scriptsize{0.1476} \\ \hline
	\end{tabular}
\end{table}

\subsection{Training Data Selection}

\begin{figure*}
	\centering
\vspace{-1.0em}
\subfloat[MNIST]{\includegraphics[width=0.34\linewidth]{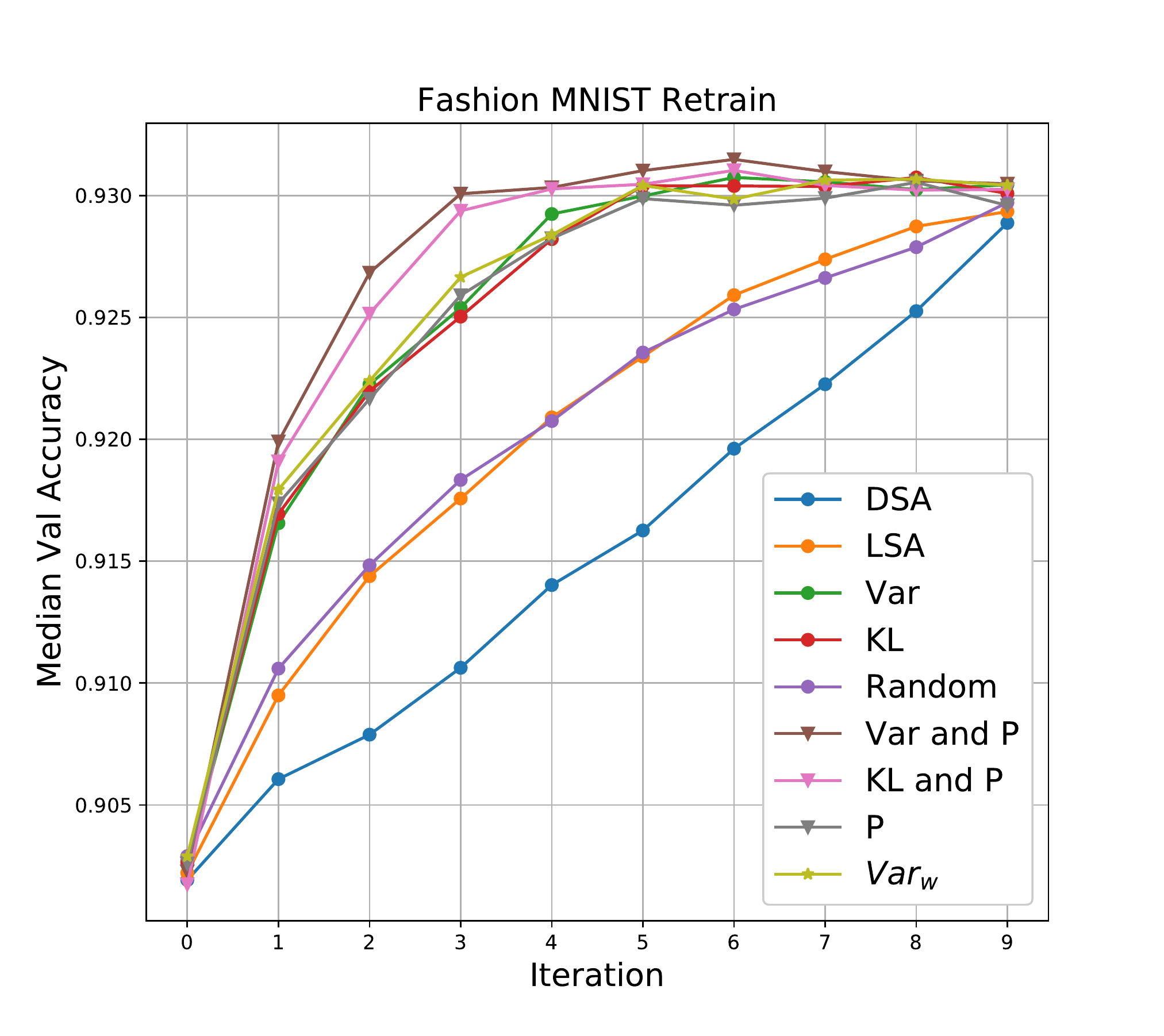}}
\vspace{-1.0em}
	\subfloat[CIFAR-10]{ \includegraphics[width=0.37\linewidth]{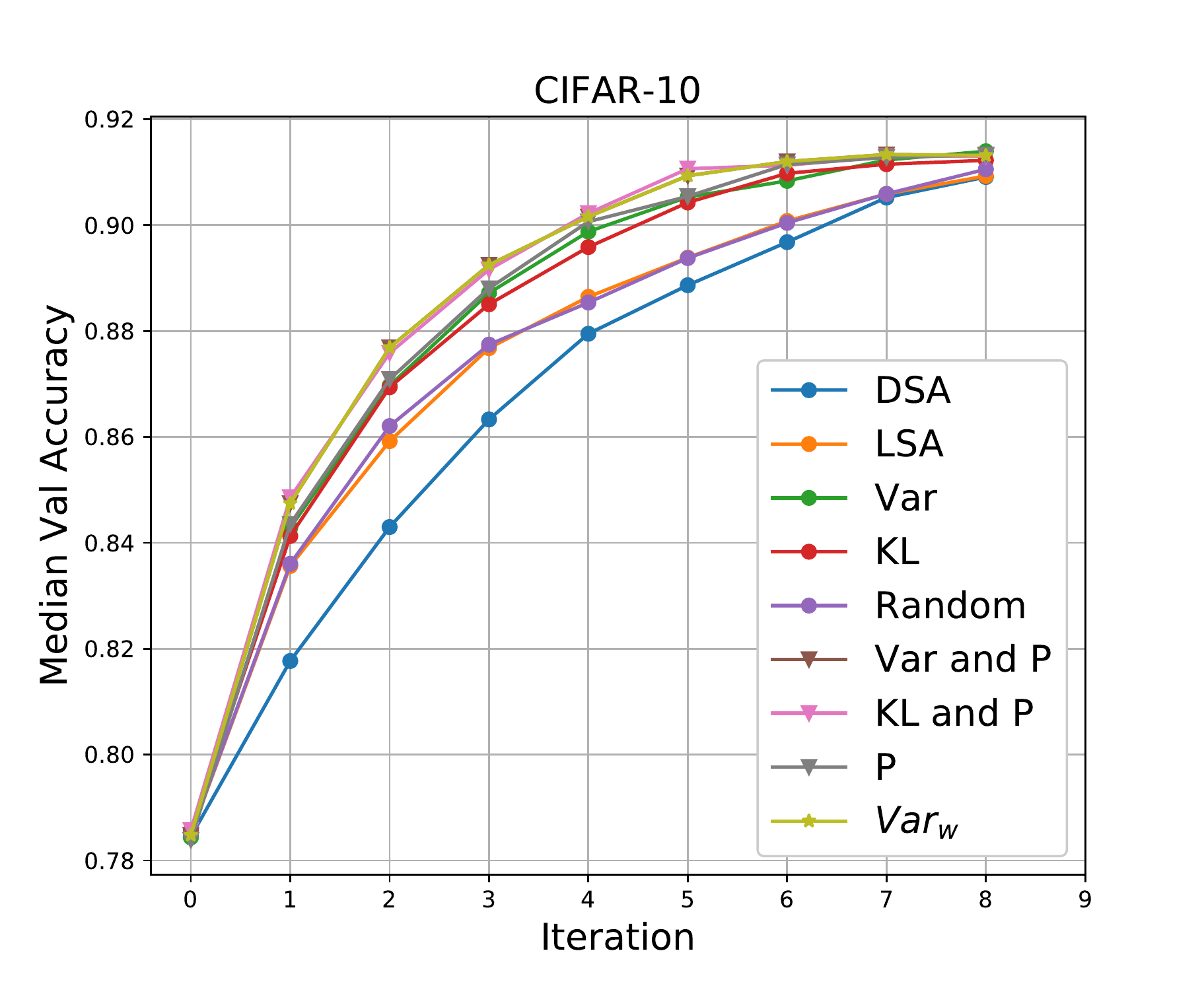}}

\caption{Validation accuracy over a fixed set of original test data and achieved by successively augmenting the training data with 5,000 data (at each iteration) selected by the different metrics. X-axis denotes the number of iterations, while Y-axis shows the median accuracy over 5 repetitions. $Var_w$ as well as $Var$ and KL combined with tie-breaking $P$ yield the best improvements, while random picking, $LSA$ and $DSA$ perform poorer.}	
\label{fig:retrain_result}
	\Description{Retrain Result}
\end{figure*}

Figure \ref{fig:retrain_result} shows the accuracy achieved by augmenting, at each iteration, the training data with 5,000 data selected according to the different metrics. We see that $Var$, $KL$ and $MaxP$ always outperform random selection, DSA and LSA. At iteration 3, the uncertainty metrics achieved a gain in accuracy (since the initial training set, i.e. iteration 0) more than 45\% higher than the other metrics (+2.2\% vs +1.5\%) on MNIST, and more than 14\% (+8\% vs +7\%) on CIFAR-10. $Var_w$ performs similarly on MNIST and better (+9\%) on CIFAR-10. 

In addition to the metrics considered so far, we augment $Var$ and KL with a tie-breaking method: when two inputs have the same $Var$ or KL scores, we prioritize the input that has the lowest $MaxP$ score. Interestingly, those two new metrics (denoted by ``\emph{Var and P}'' and ``\emph{KL and P}'') further improve the accuracy in the case of MNIST (+2.7\%) and are comparable to $Var_w$ for CIFAR-10. 

Here it must be noted that, while the raw accuracy values may seem to have small differences, they are due to the high initial accuracy of the model. Improving beyond this level is challenging. Overall, the differences in accuracy gain after 2 to 3 iterations of \emph{Var and P} and \emph{KL and P} over random selection and $LSA$, $Var$ and $KL$ are approximately 80\% on MNIST and 29\% on CIFAR-10.

\begin{figure*}
	\centering
	\includegraphics[width=0.95\linewidth]{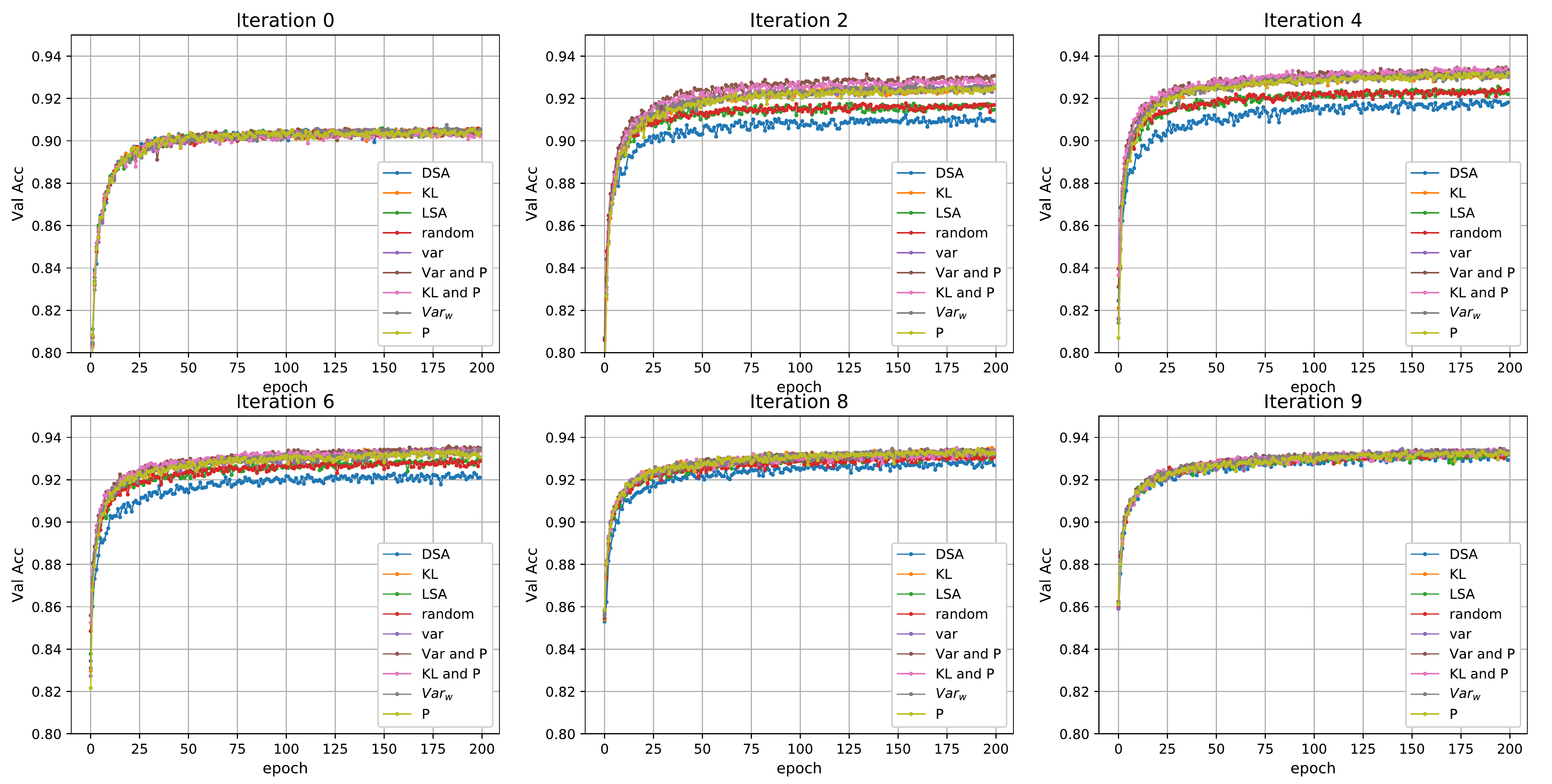}
	\caption{Validation accuracy (Y-axis) over the 200 epochs (X-axis), for all metrics and at the 1st, 2nd, 4th, 6th, 8th and 9th iterations of the retraining algorithm. At each particular iteration, the four metrics based on uncertainty make the validation accuracy converge faster than random picking, LSA and DSA, especially during the intermediate iterations.}
	\label{fig:retrain_learing_val_acc_curve}
	\Description{Retrain Result}
\end{figure*}

To further demonstrate the ability of the metrics to guide the training sample selection, Figure \ref{fig:retrain_learing_val_acc_curve} shows the accuracy achieved over the different epochs of each retraining iterations. We observe that the use of our uncertainty metrics allow the model to converge faster, especially during the intermediate iterations.

\section{Conclusion}
 In this paper, we proposed a set of test selection metrics for Deep Learning systems based on the notion of model uncertainty (estimated by neuron dropouts).  
 We showed the ability of these metrics to trigger misclassifications using both real and adversarial data and found medium to strong correlations. We also demonstrated that the metrics lead to major classification accuracy improvement (when selecting data for retraining), achieving a gain in accuracy of up to 80\% higher than the current state-of-the-art metrics and random selection. 
 
Our work forms the first step towards a long-term goal of equipping researchers and practitioners with test assessment metrics for self-learning systems. Our automatic data selection metrics pave the way for the systematic and objective selection of test data, which may lead to standardised ways of measuring test effectiveness. 



%


%
\bibliographystyle{ACM-Reference-Format}
\bibliography{sample-base,mcr}

%
\appendix

\end{document}